\documentclass[10pt]{article}

\usepackage[utf8]{inputenc} 
\usepackage[T1]{fontenc}    
\usepackage[colorlinks, linkcolor=blue, citecolor=black]{hyperref}       
\usepackage{url}            
\usepackage{booktabs}       
\usepackage{amsfonts}       
\usepackage{nicefrac}       
\usepackage{microtype}      
\usepackage{xcolor}         
\usepackage{mathtools, nccmath}
\usepackage{minitoc}
\usepackage[inline, nomargin, draft]{fixme}
\fxsetup{theme=color}
\definecolor{fxnote}{rgb}{0.858, 0.188, 0.478}

\usepackage{microtype}
\usepackage{graphicx}
\usepackage{booktabs} 

\usepackage{comment}
\usepackage{pgfplots}
\usepackage{url}

\usepackage{amsfonts}       
\usepackage{amsmath}
\usepackage{amsfonts}
\usepackage{amssymb}
\usepackage{comment}
\usepackage{pythonhighlight}

\usepackage{amsthm}
\usepackage{amscd}
\usepackage{t1enc}
\usepackage{enumerate}
\usepackage{indentfirst}
\usepackage{listings}
\usepackage{graphicx}
\usepackage{graphics}
\usepackage{pict2e}
\usepackage{epic}
\usepackage{epstopdf} 
\usepackage{listings}
\usepackage{minitoc}

\newcommand{\bE}{\mathbb{E}}

\newcommand{\bI}{\mathbb{I}}

\usepackage{amsmath}

\usepackage{empheq}

\renewcommand{\tilde}{\widetilde}
\renewcommand{\nu}{\vartheta}

\usepackage{float}
\usepackage{bm}
\usepackage{subcaption}
\usepackage{wrapfig}

\newtheorem{corollary }{Corollary }
\newtheorem*{theorem*}{Theorem}

\newtheorem{definition}{Definition}

\usepackage{apptools}
\AtAppendix{\counterwithin{lemma}{section}}
\AtAppendix{\counterwithin{theorem}{section}}
\AtAppendix{\counterwithin{corollary}{section}}
\theoremstyle{definition}
\usepackage{algorithm}
\usepackage{algorithmic}

\usepackage{environ}
\usepackage{tikz}
\usepackage{float}

\newcounter{todos}

\usepackage{epstopdf} 
\usepackage{wrapfig}
\usepackage{algorithm}
\usepackage{algorithmic}
\usepackage{amsthm}
\usepackage{tabularx}
\usepackage{float}
\usepackage{setspace}
\usepackage{fancyhdr}

\allowdisplaybreaks
\usepackage{authblk}

\usepackage[left=1in, right =1in, top=1in, bottom=1in]{geometry}
\pgfplotsset{compat=1.14}
\usepackage{pythonhighlight}
\usepackage[most]{tcolorbox}

\newtcbox{\mymath}[1][]{%
    nobeforeafter, math upper, tcbox raise base,
    enhanced, colframe=black!30!black,
    colback=white!30, boxrule=1pt,
    #1}
\usepackage{xspace}
\usepackage{adjustbox}
\newcommand{\algname}{\texttt{IntegrAI}\xspace}

\title{\textbf{Effective Human-AI Teams via Learned Natural Language Rules and Onboarding }}
\author[1,2]{Hussein Mozannar}
\author[1,2]{Jimin J Lee}
\author[1,3]{Dennis  Wei}
\author[1,3]{Prasanna  Sattigeri}
\author[1,3]{Subhro Das}
\author[1,2]{David Sontag}

\affil[1]{MIT-IBM Watson AI Lab}
\affil[2]{Massachusetts Institute of Technology}
\affil[3]{IBM Research}

\date{}
\begin{document}

%

%
\newgeometry{left=1in, right =1in, top=0.6in, bottom=0.7in}

\begin{titlepage}
\maketitle
\vspace{-4em}
\begin{abstract}
People are relying on AI agents to assist them with various tasks. The human must know when to rely on the agent, collaborate with the agent, or ignore its suggestions. In this work, we propose to learn rules, grounded in data regions and described in natural language, that illustrate how the human should collaborate with the AI. Our novel region discovery algorithm finds local regions in the data as neighborhoods in an embedding space where prior human behavior should be corrected. Each region is then described using a large language model in an iterative and contrastive procedure. We then teach these rules to the human via an onboarding stage. Through user studies on object detection and question-answering tasks, we show that our method can lead to more accurate human-AI teams. We also evaluate our region discovery and description algorithms separately.
\end{abstract}
\vspace{-1.5em}
\begin{figure*}[h]
    \centering
    \includegraphics[width=0.88\textwidth]{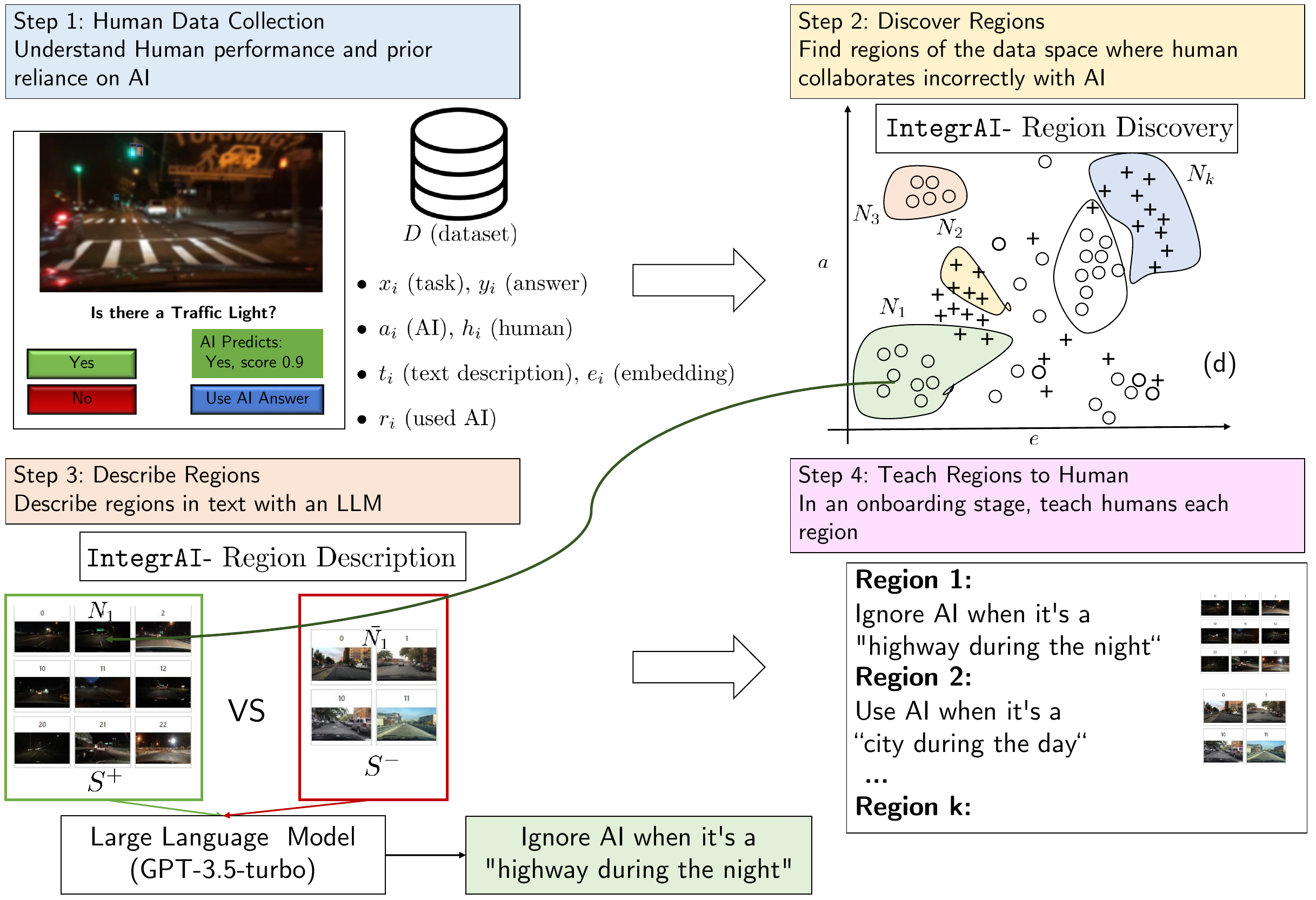}
    \caption{The proposed onboarding approach with the \algname algorithm.}
    \label{fig:overview}
\end{figure*}
\end{titlepage}

\section{Introduction}
\label{sec:intro}

As artificial intelligence (AI) becomes more ubiquitous and sophisticated, humans are increasingly working alongside AI systems to accomplish various tasks, ranging from medical diagnosis \cite{beede2020human,gaube2021ai}, content moderation \cite{gillespie2020content} to writing \cite{coenen2021wordcraft} and programming \cite{dakhel2022github}. One of the promises of AI is to enhance human performance and efficiency by providing fast and accurate solutions. However, the literature on human-AI collaboration has revealed that humans often underperform expectations when working with AI systems \cite{jacobs2021machine,liu2021understanding,kawakami2022improving,vodrahalli2022uncalibrated}. Moreover, studies have shown that providing explanations for the AI's predictions often does not yield additional performance gains and, in fact, can make things worse \cite{liu2021understanding,buccinca2020proxy,hase2020evaluating}.
These negative results of human-AI performance may be attributed to a few possible reasons. First, humans can have miscalibrated expectations about AI's ability, which leads to over-reliance \cite{bansal2019beyond}. Second, the cost of verifying the AI's answer with explanations might be too high, thus providing a bad cost-benefit tradeoff for the human \cite{vasconcelos2023explanations} and leading to bad outcomes including under-reliance on the AI \cite{nanji2014overrides}. Finally, the AI explanations do not enable the human to verify the correctness of the AI's answer and thus are not as useful for human-AI collaboration \cite{fok2023search}.

The central question remains: how can we collaborate better with AI models? In this work, we propose an intuitive framework for thinking about human-AI collaboration where the human first decides on each example whether they should rely on the AI, ignore the AI, or collaborate with the AI to solve the task. We refer to these three actions as the \emph{AI-integration decisions}. The human-AI team will perform optimally if the human knows which action is best on a task-by-task basis. We propose \algname (Figure~\ref{fig:overview}), an algorithm that leverages data from baseline human interactions with the AI to learn near-optimal integration decisions, in the form of \textit{natural language rules} that are easily understandable. These rules are then taught to the human through an onboarding stage, analogous to an onboarding class that humans might take before operating machines and equipment.
Onboarding additionally calibrates the human's expectations about AI performance. We further investigate surfacing the AI-integration decisions found by \algname as recommendations to the human within an  AI dashboard used after onboarding. The hope is that onboarding and the dashboard help the human know which action they should take, thereby leading to effective AI adoption for enhanced decision-making.

Learning AI-integration rules 
requires a dataset of paired examples and human predictions (Figure~\ref{fig:overview} Step 1). Each rule is defined as a bounded local region centered around a learned point in a potentially multi-modal embedding space spanning the task space and natural language (Figure~\ref{fig:overview} Step 2 ). For example, CLIP embeddings \cite{radford2021learning} connect image and text spaces for tasks involving images, and typical text embeddings \cite{reimers-2019-sentence-bert} are used for natural language tasks such as question answering. The regions are obtained with a novel region discovery algorithm. Then, a text description of the region is generated, resulting in a rule that indicates whether the human should ignore, rely on, or collaborate with the AI. We obtain descriptions using a novel procedure that connects the summarization ability of a large language model (LLM) \cite{brown2020language} with the retrieval ability of embedding search to find similar and dissimilar examples. The procedure first queries the LLM to describe points inside the region (Figure~\ref{fig:overview} Step 3 ). The embedding space is 
then leveraged to find counterexamples inside and outside the region to refine the description.

We first evaluate the ability of our region finding and region description algorithms to find regions that will aid the human-AI team in several real-world datasets with image and text modalities. We then investigate the efficacy of both algorithms in synthetic scenarios where we know the ground truth regions. Finally, we conduct user studies on tasks with real-world AI models to evaluate our onboarding and AI-integration recommendation methodology. Our main task is detecting traffic lights in noisy images \cite{xu2017end} from the perspective of a road car, motivated by applications to self-driving cars. The user study reveals that our methodology significantly improves the accuracy of the human-AI team by 5.2\% compared to no onboarding. We investigate a second task of multiple choice question answering using the MMLU dataset \cite{hendrycks2020measuring} and find that onboarding has no effect on performance and that only displaying AI-integration recommendations has a negative effect.
To summarize, the key contributions of this paper are as follows: 

\begin{itemize}
    \item Our region discovery algorithm finds regions that help the human know when to rely on the AI (\algname-Discover Section \ref{sec:learn_rules:discovery}).
    \item Our region description algorithm can describe regions using an LLM by contrasting points inside and outside the region.  (\algname-Describe Section \ref{sec:learn_rules:description}). We evaluate the performance of our algorithms in isolation in experiments in Section \ref{sec:method_eval}.
    \item We demonstrate the effect of onboarding and displaying AI-integration recommendations on two real-world tasks, and find that onboarding has a significant positive effect in one task whereas the integration recommendations are not useful in the second task(Section \ref{sec:studies}). In all studies, we present users with information about both human and AI performance in a Human-AI card (Section \ref{sec:onboarding}).
\end{itemize}

\section{Related Work}

This work builds on our previous work in \cite{mozannar2022teaching} where we proposed an onboarding procedure that involved participants describing different regions of the data space where AI made significant mistakes or was significantly better than human performance. We found that only 50\% of participants had accurate descriptions of the underlying regions; we could measure this percentage as the AI model was synthetic. The participants who accurately guessed the correct regions had significantly better performance than those who didn't. This motivates our work on automating the process of describing the regions and more rigorously evaluating the impact of onboarding.

A growing literature of empirical studies on AI-assisted decision making has revealed that human-AI teams do not perform better than the maximum performance of the human or AI alone even with AI explanations \cite{bansal2020does,sperrle2021survey,liu2021understanding}. This can be summarized with the following conjecture in equation form: $\mathrm{ Human +AI} \leq \max(\mathrm{Human},\mathrm{AI})$ (where accuracy is the unit). Note that equality can be achieved by an expert deferral system where a second AI model decides who between the human or the AI should predict \cite{mozannar2020consistent}.

\cite{lai2022human} proposes a method for human-AI collaboration via conditional delegation rules that the human can write down. Our framework enables the automated learning of such conditional delegation rules for more general forms of data that can also depend on the AI output. \cite{vodrahalli2022uncalibrated} proposes to modify the confidence displayed by the AI model to encourage and discourage reliance on the AI model appropriately. However, this technique deliberately misleads the human on the AI model ability. Our methodology incorporates similar ideas by learning the human prior function of reliance on the AI and then improving on it with the learned integration recommendations; however, we display these recommendations in a separate dashboard without modifying the AI model output. 
A related approach to our methodology by \cite{ma2023should} is to adaptively display or hide the AI model prediction and display the estimated confidence level of the human and the AI on a task of predicting whether a person's income exceeds a certain level. They show that displaying the confidence of the human and the AI to the human improves performance. Our method is able to learn the confidence level of the human and the AI, but also incorporates how the human utilizes the AI and describes the regions where AI vs human performance is different. \cite{cabrera2023improving} presents a similar approach to our AI recommendations, however, they use simulated and faked AI models and descriptions of behavior while we are able to obtain automated generation of these descriptions of AI behavior.

A growing literature exists on onboarding humans to work with AI models \cite{ribeiro2016should,cai2019hello,lai2020chicago,mozannar2022teaching,cabrera2023improving,kawakami2023training}. Our work differs in enabling the automated creation of onboarding material without any human in the loop.
 We compare our approach to a representative work from a body of research on discovering regions of errors in AI models \cite{eyuboglu2021domino,bharadhwaj2021auditing,ribeiro2022adaptive,wu2019errudite,zhang2022drml,wiles2022discovering,d2022spotlight,rajani2022seal,jain2022distilling}. Note however that our work focuses on regions of \emph{disparate performance} between human and AI. Learning to defer methods learn models that decide using a secondary AI model who between the human and the AI classifier should predict \cite{madras2018predict,raghu2019algorithmic,mozannar2020consistent,charusaie2022sample,mozannar2023should}, whereas \cite{lee2021fair, shah2022selective} propose methods to ensure fairness in such selective deferral settings. This paper, in contrast, focuses on the reverse setting where the human makes all decisions but we do utilize some of the thinking from that literature. Our AI-integration recommendations are also related to personalized policies \cite{bhatt2023learning}. Our MMLU experiments share similarities with recent work \cite{bowman2022measuring,zhang2023taking,tong2023mass,joshi2023machine}.
Further comparison to prior work can be found in Appendix \ref{apx:extend_related_work}.

\section{AI Assisted Decision Making}\label{sec:setting}
\begin{figure}[h]
    \centering
    \includegraphics[width=\textwidth]{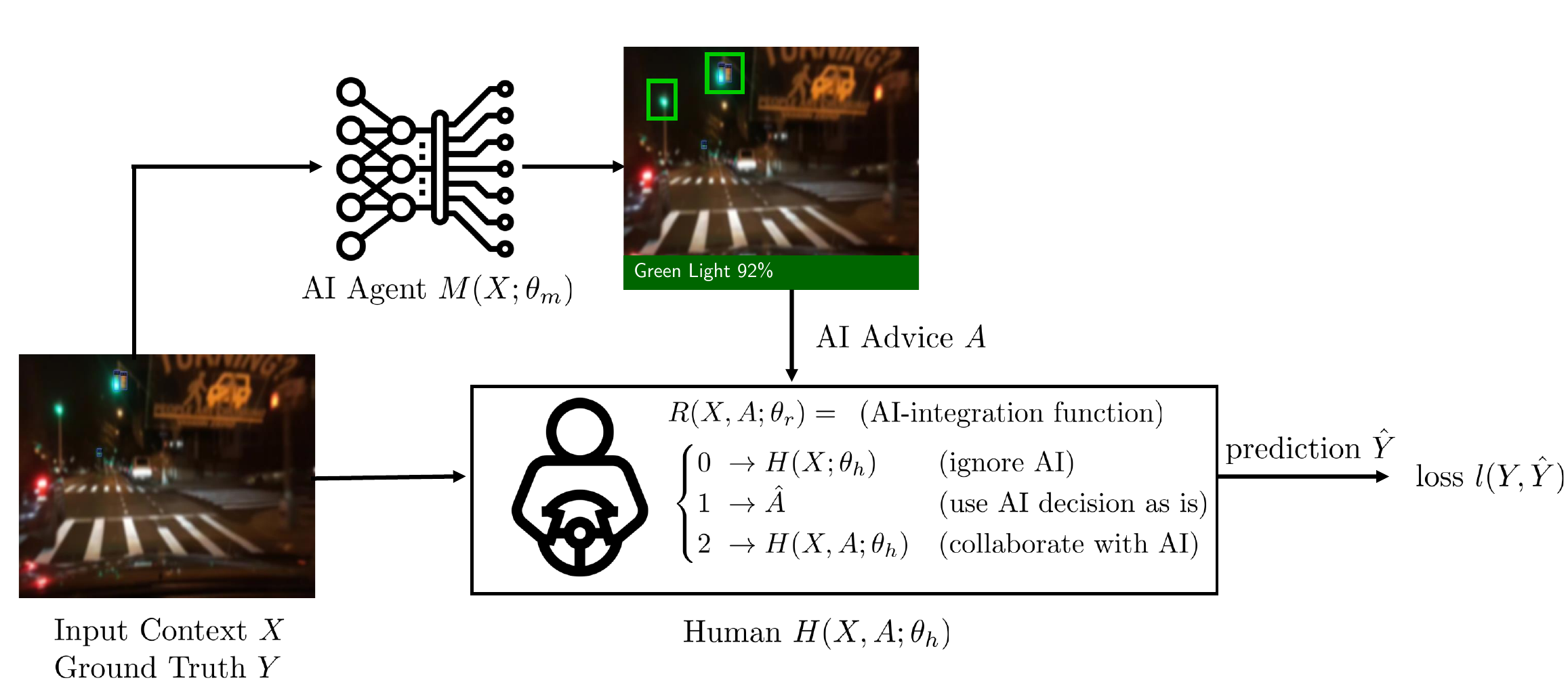}
    \caption{The setting of AI-assisted decision making studied in this work. We show an example of an AI system providing assistance to a human driver to inform them about traffic lights in a low visibility situation. The AI  provides advice to the human who then incorporates it to make a final decision.}
    \label{fig:setting_eq}
\end{figure}

\paragraph{Setting.} We consider a setting where a human is making decisions with the help of an AI agent who provides advice to complete a \textbf{task}. Formally, the \textbf{human} has to make a decision $Y \in \mathcal{Y}$ given access to information about the context as $X \in \mathcal{X}$ and the AI's advice $A \in \mathcal{A}$. We denote the human as a potentially randomized function $H(X,A;\theta_h)$ with parameters $\theta_h$ which are unobservable. On the other hand, the \textbf{AI} agent provides advice based on its viewpoint of the context $X$ according to $M(X;\theta_m):= A \in \mathcal{A}$. The advice always includes a candidate decision $\hat{A}$ and possibly an explanation of the decision. We assume that the observed tasks are drawn from an underlying distribution, $\mathbb{P}_{X,Y}$, over the contexts of AI and human, and the ground truth. The setting is illustrated in Figure \ref{fig:setting_eq}.

\paragraph{Task Metrics.} The human wants to make a decision that optimizes various metrics of interest. Given a ground truth decision $Y$ and a chosen decision $\hat{Y}$, the loss is given by $l(Y,\hat{Y}): \mathcal{Y} \times \mathcal{Y} \to \mathbb{R}^+$. In our example, this could be the 0-1 loss $\bI_{Y = \hat{y}}$. We denote the loss $L(H,M)$ of the Human-AI team over the entire domain as:
\begin{equation}
L(H,M) := \bE_{x,y \sim \mathbb{P}} \left[ l\left(y, H\left(x, M(x;\theta_m); \theta_h \right) \right) \right]
\end{equation} 
We are further interested in metrics that convey efforts undertaken by the human during the process. Particularly, we focus on \emph{time to make a decision}, which can be measured when the human makes decisions. 

\paragraph{Human-AI team.} The decision of the Human-AI team is represented by the function $H(X,A;\theta_h)$. The human without the AI is denoted by $H(X,\emptyset;\theta_h):=H(X;\theta_h)$, which is obtained by setting the AI advice to the empty set, i.e. no advice. In theory, we expect the human with advice to perform at least as well as without advice, simply because the human can always ignore the advice. However, the literature on Human-AI teams has clearly demonstrated that this is often not the case \cite{lai2021towards}. An \textbf{effective} Human-AI team is one where the human with the AI's advice achieves a more favorable trade-off in terms of the metrics than without the advice.

\paragraph{Framework for Cooperation.} Our insight into forming an effective team is to explicitly recommend when the human should consider the advice and how to incorporate it into their decision. We propose a two-stage framework for the human to cooperate with the AI: the human first decides whether to ignore the AI advice, use the AI's decision or integrate the AI advice to make a decision with explicit cooperation. Each of these three cases provides a clear path to the second stage of making the final output decision.
\begin{definition}
The  AI-integration function $R(X,A;\theta_r)$, also referred to as integrator, formalizes a framework for the human to cooperate with the AI:
\begin{equation}
R(X,A;\theta_r) = \begin{cases}
0 \ \to H(X;\theta_h) \ \ &\textrm{(ignore AI)} \\
1 \ \to \hat{A} \ \ &\textrm{(use AI decision as is)} \\
2 \ \to H(X,A;\theta_h) \ \ &\textrm{(collaborate with AI)}
\end{cases}
\end{equation}
In this work, we only consider the actions of ignoring or using the AI decision, $R \in \{0,1\}$, and leave the action of $R=2$ for future work.
\end{definition} 
The integration function can be thought of as a specific formalization of the human mental model of the AI \cite{bansal2019beyond,bansal2019updates}. 
Given an integration function $R$ and a human $H$, we can define a hypothetical human decision maker $H_R$ who first computes $R$ and then follows its recommendation to make a final decision. Similarly, for each human $H(X,A;\theta_h)$, we can associate an integration function $R_H$, such that $H_{R_H} = H$. By fixing $H(X;\theta_h)$, one can try to minimize the loss $L(H_R,M)$ over all possible choices of $R$, an optimal point of such an integration function is denoted $R^*$. Following the recommendations of the optimal AI-integration function leads to an \textbf{effective} Human-AI team. 

\paragraph{Learning Rules and Onboarding.} 
There are two problems that need to be solved to achieve this vision: how do we learn such an $R^*$ and how can we ensure that the human follows the recommendations of this $R^*$? In the next section, we outline how we approximate the optimal integration function; this is fundamentally a machine-learning problem with its own challenges that we tackle in  Section~\ref{sec:learn_rules}. The second obstacle is that the human should know to follow the recommendations of $R^*$.  To ensure this, we propose an \textbf{onboarding} stage where the human learns about the AI and the optimal integration function. We additionally propose displaying the recommendations as part of an AI dashboard. In the onboarding stage, we will help the human shape their internal parameters $(\theta_h, \theta_r)$ to improve performance. This is a human-computer interaction (HCI) problem that we tackle in Section~\ref{sec:onboarding}. 

\section{Learning Rules for Human-AI Cooperation: \algname}\label{sec:learn_rules}

In this section, we discuss how to learn an \emph{integrator} function $\hat{R}: \mathcal{X} \times \mathcal{A} \to \{0,1\}$ to approximate an optimal integrator while being understandable to the human. We first describe the ingredients for this learning (integrator as a set of regions, objective function, dataset) before detailing how we learn regions and describe them in Sections~\ref{sec:learn_rules:discovery} and \ref{sec:learn_rules:description} respectively.

\paragraph{Integrator as a Set of Regions.} Since the integrator $\hat{R}$ will be used to both onboard the human and provide test-time recommendations as part of an AI dashboard, it should be easily understandable to humans. If the goal was simply to build the most accurate integrator, we could use work on learning to defer \cite{mozannar2020consistent,charusaie2022sample,mozannar2023should}.
To address the requirement of understandability, we propose to parameterize the integrator in terms of a set of local data regions, each with its own integration decision label as well as a natural language description.
More specifically, we aim to learn a variable number of regions $N_1, N_2, \cdots$ as functions of $(X,A)$, the observable context and the AI's advice. Each region $N_k$ consists of the following: 1) an indicator function $I_{N_k}: \mathcal{X} \times \mathcal{A} \to \{0,1\}$ that indicates membership in the region, 
2) a textual description of the region $T_k$, and 3) a takeaway for the region consisting of an integration decision $r(N_k) \in \{0,1\}$.
We additionally want these regions to satisfy a set of constraints so that they are informative for the human and suitable for onboarding.

\paragraph{Maximizing Human's Performance Gain.} Since we are working with human decision-makers, we have to account for the fact that the human implicitly has a \textbf{prior} integration function $R_0$, which represents how the human would act without onboarding.  Thus, in learning integrator regions, our goal is to maximize the human performance gain relative to their prior.
The performance gain is defined as follows for points in a region $N$:
\begin{equation} \label{eq:gain_G}
G(N, \hat{R}, R_0) = \sum_{i \in N} l(y_i,H_{R_0}(x_i,a_i)) - l(y_i,H_{\hat{R}}(x_i,a_i)), 
\end{equation}
where $l$ is the loss defined in Section~\ref{sec:setting}. 
Note that the notion of a human's prior mental model was also discussed by \cite{mozannar2022teaching} but we expand on the notion and are able to learn priors as we discuss below. 

\paragraph{Dataset with Human Decisions.} We assume we have access to a 
dataset $D=\{x_i,y_i,h_i,r_{0i}\}_{i=1}^n$ sampled from $\mathbb{P}$, where $x_i$ is the AI-observable context, $y_i$ is the optimal decision on example $i$, $h_i$ is a human-only decision, defined in Section~\ref{sec:setting} as $H(x_i;\theta_h)$, and $r_{0i} \in \{0,1\}$ is an indicator of whether the human relied on AI on example $i$. 
We thus regard the samples $\{r_{0i}\}_{i=1}^n$ as a proxy for prior human integration function $R_0$. 
The prior integration decisions of the human $r_{0i}$  are collected through a data collection study where the human predicts with the AI without onboarding. For example in Figure \ref{fig:interface}, when the human presses on the "Use AI" button we record $r_{0i}=1$, and 0 otherwise. The human predictions $h_{i}$ are collected through a second data collection study where the human makes predictions without the AI.  
We also assume that we are given an AI model $M$, from which we obtain AI decisions $\hat{a}_i \in a_i$ from the AI advice $a_i = M(x_i; \theta_m)$ \footnote{Moreover, we assume the AI was not trained on the dataset $D$ so that we can use $D$ to obtain an unbiased measurement of AI performance; this is crucial, as otherwise, we might overestimate its performance. 
}. 
Given the dataset $D$, AI decisions $\hat{a}_i$, and loss $l(.,.)$, we can define optimal per-example integration decisions $r^*_i$ by comparing human and AI losses on the example: 
$r^*_i = \bI\left(l(y_i,h_i) > l(y_i,\hat{a}_i)\right)$.

\subsection{Region Discovery Algorithm}
\label{sec:learn_rules:discovery}

In this subsection, we describe a sequential algorithm that starts with the prior integration function $R_0$ and adds regions $N_k$ one at a time. 
\paragraph{Representation Space.} The domain $\mathcal{X}$ for the task may consist of images, natural language, or other modalities that possess an interpretable representation space. We follow a similar procedure for all domains. The first step is to map the domain onto a potentially cross-modal embedding space using a mapper $E(.)$, where one of the modes is natural language. The motivation is that such an embedding space will have local regions that share similar natural language descriptions, enabling us to learn understandable rules. For example, for natural images, we use embeddings from the CLIP model \cite{radford2021learning}. 
 The result of this step is to transform the dataset $\{x_i\}_{i=1}^n$ into a dataset of embeddings $\{e_i\}_{i=1}^n$ where $e_i \in \mathcal{E_X} \subseteq \mathbb{R}^d$. 
 
\paragraph{Region Parameterization.} We define region $N_k$ in terms of a centroid point $c_k$, a scaled Euclidean neighborhood around it, and an integration label $r_k \in \{0,1\}$. The neighborhood is in turn defined by a radius $\gamma_k \in \mathbb{R}$ and element-wise scaling vector $w_k$. Both $c_k$ and $w_k$ are in $\mathcal{E_X} \times \mathcal{A}$, the concatenation of the embedding space and the AI advice space. The indicator of belonging to the region is then $I_{N_k}(e_i,a_i) =  \bI_{ ||w_k \circ ((e_i,a_i)-c_k)||_2 < \gamma_k}$, where $\circ$ is the Hadamard (element-wise) product. 

\paragraph{Region Constraints.} 
We add the following constraints on each region $N_k$ to make them useful to the human during onboarding. First, the region size in terms of fraction of points contained must be bounded from below by $\beta_l$ and above by $\beta_u$. Second, the examples in each region must have high agreement in terms of their optimal per-example 
integration decisions $r^*_i$. Specifically, at least a fraction $\alpha$ of the points in a region must have the same value of $r^*_i$. Finally, each region must have at least a gain \eqref{eq:gain_G} of $\delta$, simply speaking adding the region to our integrator provides a gain of $\delta$ in terms of our loss $l$.

\paragraph{\algname-Discover.} Our procedure is fully described in Algorithm \ref{alg:algorithm_disover}.  In round $k$, we add a $k$th region to the integrator $R_{k-1}$ from the previous round (with $k-1$ regions) to yield $R_k$. 
After $T$ rounds, the updated integrator $R_T$ is defined as follows: Given a point $(x,e,a)$ where $e$ is the embedding of $x$, if it does not belong to any of the regions $N_1,\dots,N_T$, then we fall back on the prior. Otherwise, we take a majority vote over all regions to which $(x,e,a)$ belongs:
\begin{equation} \label{eq:updated_integrator}
    R_{T}(x,e,a) = \begin{cases}
        R_0(x,a) \quad  &\text{if } I_{N_k}(e,a) = 0, \; k=1,\dots,T\\
        \mathrm{majority}\left(\{r(N_k): k \text{ s.t. } I_{N_k}(e,a)=1 \}\right) \quad  &\text{otherwise}. \\
    \end{cases}
\end{equation}

 In round $k$, we compute the potential performance gain for each point if we were to take decision $r$ on the point as $g_{i,r} = l(y_i, H_{R_{k-1}}(x_i,a_i)) - l(y_i, r)$; this vector of gains is denoted by $\mathbf{g}$. 
The optimization problem to find the optimal regions is non-differentiable due to the discontinuous nature of the region indicators. 
To make this optimization problem differentiable, we relax the constraints to be penalties with a multiplier $\lambda$ and replace the indicators with sigmoids scaled by a large constant $C_1$. To find a new region $R$ given a gain vector $\mathbf{g}$ we need to solve the following optimization problem:
\begin{align} \label{eq:objctive_j}
 & \max_{c, \gamma, w, r} \ \ J(c,\gamma,w,r;\mathbf{g}) :=  \sum_{i=1}^n \sigma(C_1(-||w \circ ((e_i,a_i)-c)|| + \gamma))  \cdot g_{i,r} \quad \textrm{(maximize gain)} \\&- \lambda \max ( \sum_{i=1}^n \sigma(C_1(-||w \circ ((e_i,a_i)-c)|| + \gamma))   \cdot  \bI_{r^*_i = r}  + \alpha n)  ,0) \quad \textrm{(consistent takeaway)}  \\ 
    &- \lambda \max\left( \sum_{i=1}^n \sigma(C_1(-||w \circ ((e_i,a_i)-c)|| + \gamma))  - \beta_u n ,0\right)  \quad \textrm{(region max size)}  \\
    & -  \lambda \max\left( -\sum_{i=1}^n \sigma(C_1(-||w \circ ((e_i,a_i)-c)|| + \gamma))  + \beta_l n ,0\right) \quad \textrm{(region min size)} 
\end{align}
The optimization variables $c, \gamma, w,$ and $r$  to find the region are all real-valued except for $r$ which is a binary variable. Thus to optimize the objective $J$, we use AdamW to optimize over $(c,\gamma,w)$ twice: once for $r=0$ and once for $r=1$ and pick the solution that has the highest objective value between the two.
The hyperparameters of the algorithm are the minimum size of the region $\beta_l$, maximum size of the region $\beta_u$, consistency of the region $\alpha$ , and minimum gain of the region $\delta$.
Further details can be found in Appendix \ref{apx:region_finding}.

\begin{algorithm}[h]
\caption{\algname-Discover}
\label{alg:algorithm_disover}
\textbf{Input}: Dataset $D$, prior integrator $R_0$, maximum number of regions $T$ 

\begin{algorithmic}[1] 
\STATE $\mathcal{N} \gets \emptyset$ (regions found so far)
\STATE $C^0=\{c^0_1,\cdots,c^0_{100}\}$: set of possible initializations from \textbf{K-medoids} on $D$ 
\FOR{$k=1,\cdots,2T$}
\FOR{$r \in \{0,1\}$}
\STATE \textbf{Compute Gain Vector:} $\mathbf{g}$ (gain from action $r$ on each point given $R_{k-1}$)
\STATE \textbf{Initialization:} $c^0, \gamma^0, w^0 \ = \  \arg\max_{c \in C^0, \gamma, w} J(c,\gamma,w,r;\mathbf{g})$ (search over $C^0$ with 200 epochs \eqref{eq:objctive_j})
\STATE \textbf{Full Optimization:} $c, \gamma, w \ = \  \arg\max_{c, \gamma, w, r} J(c,\gamma,w,r;\mathbf{g})$ (optimization starts from initialization above for 2000 epochs)
\STATE \textbf{Form candidate region:} $N_k^r$ (from $c, \gamma, w,r$)
\ENDFOR
\STATE \textbf{Form potential integrators:} $\hat{R}_0,\hat{R}_1$ by adding $N_k^0$ and $N_k^1$ respectively
\STATE $r^* = \arg\max_{r\in \{0,1\}} G(N_k^r,\hat{R}_r,R_{k-1})$ \textbf{(find best takeaway decision from last step following \eqref{eq:gain_G})}
\IF{$G(N_k^{r^*},\hat{R}_{r^*},R_{k-1}) \geq \delta$}
\STATE   $\mathcal{N} = \mathcal{N} \cup N_k^{r^*}$ (\textbf{add region to set }- as it has high enough gain)
\STATE \textbf{Form new integrator:} $R_k$ by adding region $N_k^{r^*}$ following \eqref{eq:updated_integrator}.
\ELSE
\STATE \textbf{Form  integrator:} $R_k \gets R_{k-1}$ (no update) 
\ENDIF
\IF{$|\mathcal{N}|$==T}
\STATE \textbf{break} (exit for loop) - we have enough regions
\ENDIF
\ENDFOR
\end{algorithmic}
\textbf{Return:} Set of  Regions discovered $\mathcal{N}$
\end{algorithm}

\subsection{Region Description Algorithm}
\label{sec:learn_rules:description}

\begin{figure}
    \centering
    \includegraphics[width=\textwidth]{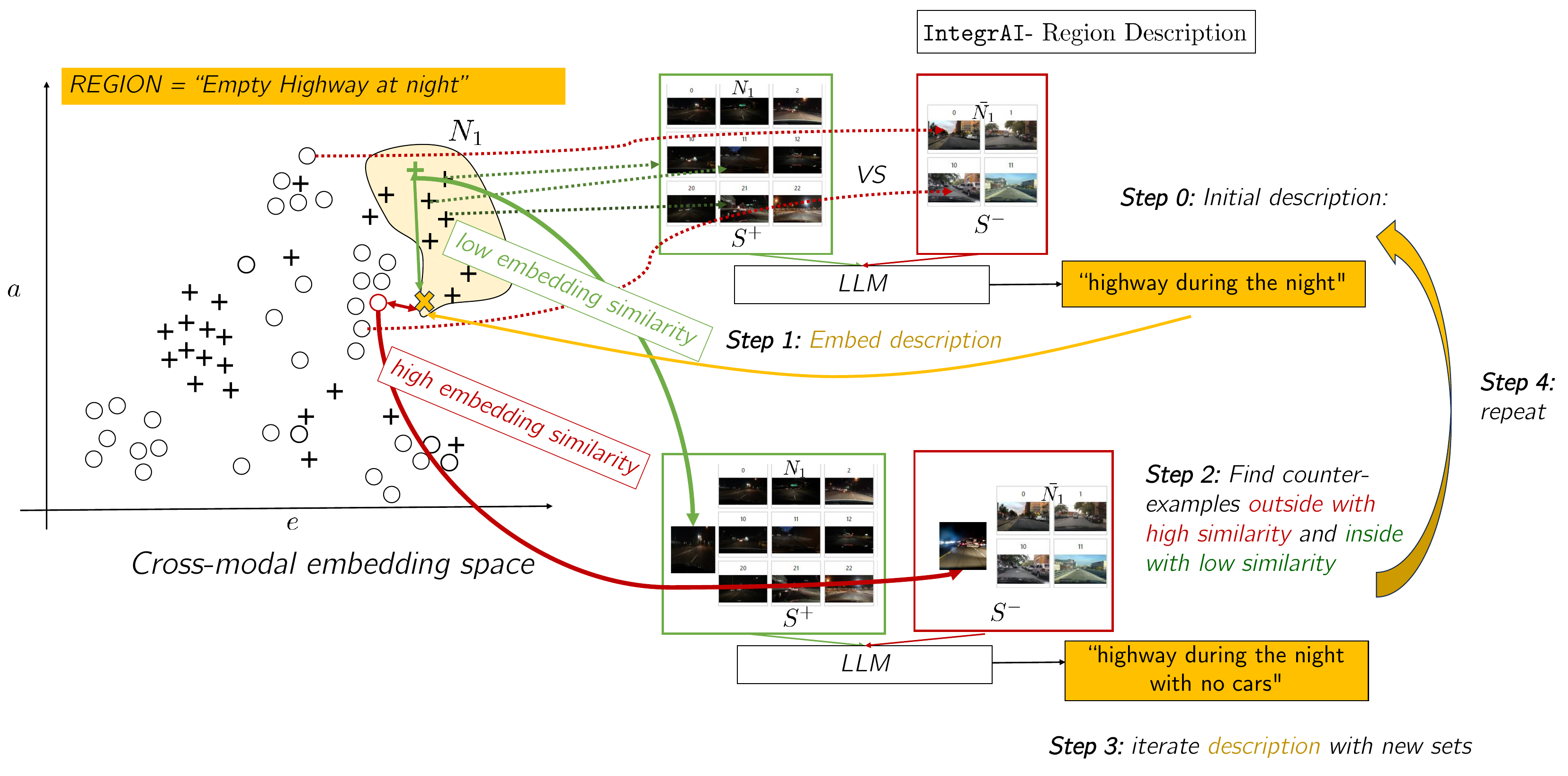}
    \caption{The \algname-Describe algorithm illustrated. To obtain a description for a region of points, the algorithm first samples a set of points inside and outside the region and gets a description from an LLM that contrasts inside versus outside (Step 0). We then embed the obtained description in our cross-modal embedding space (Step 1) and find counterexamples to that description, both points outside the region with high similarity to the description and points inside the region with low similarity to the description (Step 3). The process is repeated for as many rounds as necessary (Step 4). }
    \label{fig:describe_fig}
\end{figure}

\begin{algorithm}[h]
\caption{\algname-Describe}
\label{alg:describe}
\textbf{Input}: Dataset $D$, region $N_k$
\begin{algorithmic}[1] 
\STATE $S^+ \gets$ 15 random examples from $N_k$, $S^- \gets$ 5 random examples outside $N_k$
\STATE \textbf{Initial Region Description:} $T_{k}^0 \gets \hat{O}(S^+,S^-)$ (LLM call)
\FOR{$i=1,\cdots,m$}
\STATE \textbf{Find Counterexample outside region.} $s^-= \arg \max_{j \notin N_k} \textrm{sim}(E(T_{k}^i),e_j)$ (most similar outside region)
\STATE \textbf{Find Counterexample inside region.} $s^+= \arg \min_{j \in N_k} \textrm{sim}(E(T_{k}^i),e_j)$ (least similar inside region)
\STATE \textbf{Update Inside and Outside Sets. } $S^- \gets S^- \cup t_{s^-}$ and $S^+ \gets S^+ \cup t_{s^+}$
\STATE \textbf{Get New Region Description:} $T_{k}^i \gets \hat{O}(S^+,S^-)$ (LLM call)
\ENDFOR
\end{algorithmic}
\textbf{Return:} Region description $T_k^m$
\end{algorithm}
We now describe our region description algorithm aimed at making the rules for integration human-understandable. Natural language descriptions are a good match for this objective. Specifically, we would like to find a contrastive textual description $T_k$ of each region $N_k$ that describes it in a way to distinguish it from the rest of the data space. The algorithm is formally described in Algorithm \ref{alg:describe} and illustrated in Figure \ref{fig:describe_fig}.

\paragraph{Textual Descriptions for Regions:} The first step is to have a textual description $t_i$ for each example in our dataset $D$ based on $x_i$. If textual descriptions are not available, we can obtain them by utilizing models that map from the domain $\mathcal{X}$ to natural language, such as captioning models for images, summarization models for text, or exploiting metadata to construct a natural language description. To obtain a region description, one idea is to ask an LLM (such as GPT-3.5) to summarize all textual descriptions of points inside the region. However, there are two issues with this approach: first, the region may contain thousands of examples so we need an effective way to select which points to include, and second, the obtained region description may not contrast with points outside the region. To resolve these issues, we propose an algorithm that iteratively refines region descriptions with repeated calls to an LLM ($\hat{O}$) in Algorithm
\ref{alg:describe} (\algname-Describe).  The algorithm starts with an initial description and then at each round finds two types of counterexamples to that description: examples outside the region (type $-$) with high cosine similarity in terms of embeddings (sim) to the region description, and examples inside the region (type $+$) with low similarity to the region description. Then we add those counterexamples to our example sets and derive a new region description. We use a specially created prompt with an exemplar to the LLM to get the region description at each round; this prompt can be found in Appendix \ref{apx:region_describe}. 

\paragraph{Illustrative Example.} Suppose we want to describe a region consisting of images of highways during the night, with no cars present (see Figure \ref{fig:describe_fig} for images of BDD \cite{xu2017end}). Our method's initial description is ``The highway during the night with clear, rainy or snowy weather,'' not mentioning that the highway has no cars, particularly because the captions of examples $t_i$ only mention the presence of cars and not their absence. In the second round, the algorithm finds the counterexample $s^-$ with caption ``city street during the night with clear weather with a lot cars'' and counterexample $s^+$ ``highway during the night with clear weather.'' The new description $T_k^1$ becomes ``clear highway during the night with various weather conditions, while outside the region are busy city street at night with clear weather.'' After one more round, the description $T_k^2$ becomes ``driving on a clear and uncongested highway during the night in various weather conditions.'' A simplified version of this example is shown in Figure \ref{fig:describe_fig}. We now proceed in the next section to describe how we onboard the human decision maker using the regions.  

\section{Onboarding and Recommendations to Promote Rules}\label{sec:onboarding}

Once rules for integration have been learned as described in the previous section, the task is to teach these rules to the human with an \textbf{onboarding} stage and encourage their use at test time. We accomplish this through an \textbf{onboarding} process followed by test-time \textbf{recommendations}, as described next.

\paragraph{Human-AI Card.} 
The onboarding process accordingly consists of an \emph{introductory} phase and a \emph{teaching} phase. 
In the \emph{introductory phase}, the user is first asked to complete a few practice examples of the task on their own to gain familiarity. The user is then presented with general information about the AI model in the form of a \textbf{human-AI card}, very similar to a  model card \cite{mitchell2019model} and inspired by \cite{cai2019hello}. The card structure showcased in Table \ref{tab:information_about_ai} includes the AI model inputs and outputs, training data,  training objectives, overall human and AI performance along with AI performance on subgroups where performance deviates from average performance. This card represents the bare minimum that humans should know about the AI model before collaborating with it. The instantiation of the Human-AI card for the BDD user study is shown in Table \ref{tab:bdd_card}.

\begin{table}[H]
    \centering
        \caption{Human-AI Card presented to the human as part of onboarding}

    \begin{tabular}{lp{8cm}}
    \toprule
    Information & Description \\ \midrule
       AI Input  & What the AI uses to make its prediction  \\ 
        AI Output & What the AI provides as output (predictions, explanations, ...) \\ 
         Source of Training Data for AI & Description of data used to train the AI \\ 
         Source of Pre-Training Data of AI& Description of pre-training data that AI is based on   \\ 
        Training Objective of AI & What the AI is trying to achieve (minimize classification error, detect objects, next word prediction) \\Average AI Performance &  Relevant metrics of overall AI performance (accuracy, FPR, ...) \\ 
        Average Human Performance & Relevant metrics of overall human performance (accuracy, FPR, ...)  \\ 
        AI vs Human Performance on Subgroups &  \\ 
        \bottomrule
    \end{tabular}
    \label{tab:information_about_ai}
\end{table}

The \textbf{teaching phase} aims to provide a more detailed picture of how the human should collaborate with the AI. It is structured as a sequence of ``lessons'', each corresponding to a region $N_k$ resulting from the algorithm of Section~\ref{sec:learn_rules:discovery}. 
The specific steps in each lesson are as follows:
\begin{itemize}
    \item 
\textbf{Step 1: Human predicts on example.} A representative from the region is selected at random that has an optimal integration decision identical to that of the region. The human is asked to perform the task for the chosen representative and is shown the AI's output along with the option to use the AI's response.

\item \textbf{Step 2: Human receives feedback.} After submitting a response, the user is told whether the response is correct and whether the AI is correct.
    
\item \textbf{Step 3: From example to region learning.} The user is informed that the representative belongs to a larger region $N_k$ and is provided with the associated recommendation $r_k$, textual description $t_k$, and  AI and human performance in the region as well as the raw examples from the region in a gallery viewer. 
\end{itemize}

After completing all lessons as above, a second pass is done where the user is re-shown all lessons for which their response was incorrect. This serves to reinforce these lessons and is similar to online learning applications such as Duolingo for language learning \cite{portnoff2021methods}.
Our teaching approach is motivated by literature showing that humans learn through examples and employ a nearest neighbor type of mechanism to make decisions on future examples \cite{bornstein2017reminders,richler2014visual,giguere2013limits}. 
We follow this literature by showing concrete data examples in the belief that it effectively teaches humans how to interact with AI.
We improve on the approach in \cite{mozannar2022teaching} by incorporating an introductory phase with the Human-AI card, showing pre-defined region descriptions for each region, and iterating on misclassified examples in the teaching phase.

\paragraph{Recommendations in AI Dashboard.} At test time, we can check whether an example $x$ and its corresponding AI output $a$ fall into one of the learned regions $N_k$. If they do, then our dashboard shows the associated recommendation $r_k$ and description $t_k$ alongside the AI output $a$. 

\begin{table}[H]
\centering
\caption{The exact Human-AI card used in the user study for BDD.}

\begin{tabular}{lp{8cm}}
\toprule
\textbf{Attribute}               & \textbf{Description} \\ \midrule
Average AI Accuracy              & 78\% \\
Average Human Accuracy           & 72\% \\
AI Model Input                   & Blurry Image \\
AI Model Output                  & Prediction of traffic light, bounding box on image showing its location and a score indicating its confidence \\
Source of Training Data          & Dataset of road images from New York and San Francisco Bay Area \\
AI Training Objective            & Detect traffic lights and other objects in image \\
\bottomrule
\end{tabular}

\begin{tabular}{ll}
\toprule
\textbf{Category}                        & \textbf{Accuracy} \\
\midrule
more than 5 traffic lights in image      & 90\% \\
1 to 4 traffic lights        & 62\% \\
no traffic lights            & 86\% \\
no cars                                  & 83\% \\
daytime or overcast weather              & 75\% \\
few pedestrians                          & 76\% \\
\bottomrule
\end{tabular}

\label{tab:bdd_card}
\end{table}

\section{Method Evaluation}\label{sec:method_eval}

\paragraph{Objective.} In this experimental section\footnote{Code is available in \url{https://github.com/clinicalml/onboarding_human_ai}.}, we evaluate the ability of our algorithms to achieve three aims: (Aim~1) Learn an integration function that leads to a human-AI team with low error; (Aim~2) discover regions of the data space that correspond to the underlying regions where human vs AI performance is different; and, for our region description algorithm, (Aim~3) come up with accurate descriptions of the underlying regions. Full experimental details are in Appendix \ref{apx:method_eval}. 

\begin{wrapfigure}[19]{R}{0.45\textwidth}
    \centering
    \resizebox{0.45\textwidth}{!}{
    \includegraphics[scale=0.6]{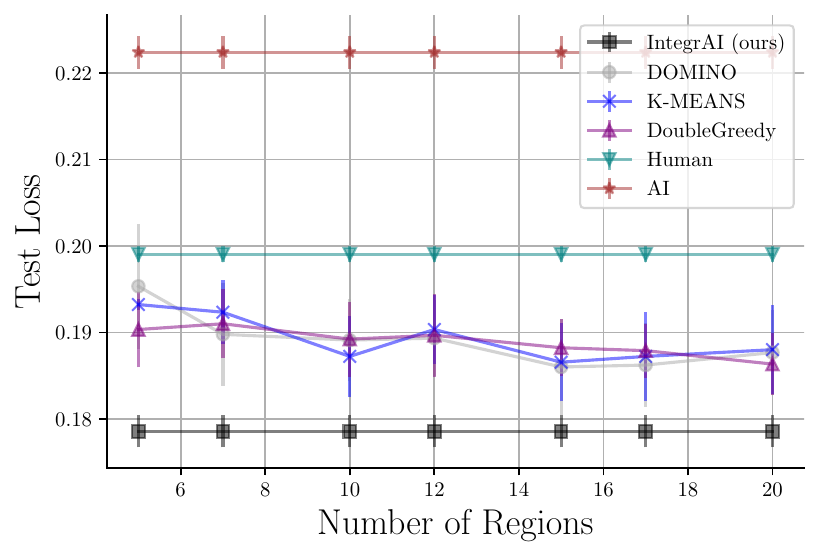}
    }
    \caption{Test Error ($\downarrow$) of the human-AI system when following the decisions of the different integrators as we vary the number of regions maximally allowed for each integrator on the BDD dataset.  }
    \label{fig:bdd_real_plot}
\end{wrapfigure}

\paragraph{Datasets and AI Models.} The experiments are performed on two image object detection datasets and two text datasets. The image datasets include Berkeley Deep Drive (BDD) \cite{xu2017end} where the task is to detect the presence of traffic lights from blurred images of the validation dataset (10k), and the validation set of MS-COCO (5k) where the task is to detect whether a person is present in the image \footnote{We extend this to detecting the presence of any object.} \cite{lin2014microsoft}. The text-based validation datasets are Massive Multi-task Language Understanding (MMLU) \cite{hendrycks2020measuring}, and Dynamic Sentiment Analysis Dataset (DynaSent) \cite{potts2020dynasent}. 
The pre-trained Faster R-CNN models \cite{ren2015faster} are considered for BDD and MS-COCO. For MMLU, a pre-trained flan-t5 model \cite{chung2022scaling} is utilized, whereas a pre-trained sentiment analysis model roBERTa-base is used for DynaSent \cite{barbieri2020tweeteval}. Each dataset is split into a 70-30 ratio for training and testing five different times so as to obtain error bars of predictions. We obtain embeddings using a sentence transformer \cite{reimers-2019-sentence-bert} for the text datasets and CLIP for the image datasets \cite{radford2021learning}

\paragraph{Baselines.} We benchmark our algorithm with different baseline methods that can find regions of the space. The baselines include: (a) DOMINO \cite{eyuboglu2021domino} which is a slice-discovery method for AI errors, (b) K-Means following the approach of \cite{rajani2022seal}, and (c) the double-greedy algorithm from \cite{mozannar2022teaching} that finds regions for Human-AI onboarding. For the regions obtained from these baselines, we compute the optimal integration decision that results in minimal loss. For our method, we set $\beta_u=0.5,\beta_l=0.01,\alpha=0.0$ for Aim 1 and  $\beta_u=0.2,\beta_l=0.01,\alpha=0.8$ for Aim 2, random prior decisions (50-50 for 0 and 1), and $\delta=2$.
In the context of region-description algorithms, we compare to the SEAL approach \cite{rajani2022seal}, a simple baseline that picks the best representative description from the existing dataset (best-caption), and ablations of our method. For Aim1 and Aim2, we repeat each experiment 5 times and report the average value and standard error (standard deviation divided by $\sqrt{5}$).

\begin{table}[htbp]
  \centering
  \begin{minipage}[b]{0.52\linewidth}
    \centering
        \caption{ Error ($\downarrow$) on the test set (in \%) of the human-AI system when following integrators resulting from different region discovery methods with $10$ regions on the different non-synthetic datasets. }

        \begin{adjustbox}{max width=1.0\textwidth}
          \label{tab:train_error_real}
\begin{tabular}{lcccc}
\toprule
 & BDD &  MMLU & DynaSent & MS-COCO  \\
\midrule
IntegrAI (ours) &\textbf{17.8 $\pm$ 0.2} &\textbf{ 45.3 $\pm$ 0.3 }& 20.2 $\pm$ 0.3 &\textbf{22.6  $\pm$ 0.4} \\
DOMINO \cite{eyuboglu2021domino}&18.9 $\pm$ 0.4 & 48.1 $\pm$ 0.2 & 20.0 $\pm$ 0.2 & 22.7 $\pm$ 0.4 \\
K-MEANS \cite{rajani2022seal} & 19.0 $\pm$ 0.5 & \textbf{45.3 $\pm$ 0.3} & 20.0 $\pm$ 0.2 &  23.2 $\pm$ 0.1\\
DoubleGreedy \cite{mozannar2022teaching} &18.9 $\pm$ 0.1 & 46.1 $\pm$ 0.6 &20.0 $\pm$ 0.2 &23.8 $\pm$ 0.4 \\
\bottomrule
\end{tabular}
  \end{adjustbox}
  \end{minipage}
  \hfill
  \begin{minipage}[b]{0.45\linewidth}
    \centering
        \caption{Clustering metrics (Adjusted Rand index \cite{santos2009use} $\uparrow$, Fowlkes–Mallows index \cite{fowlkes1983method} $\uparrow$)  of the regions (10 regions) found by the different methods on the synthetic dataset setup. }

\begin{adjustbox}{max width=1.0\textwidth}
          \label{tab:clust_synth}
\begin{tabular}{lcccc}
\toprule
 & BDD &  MMLU  & MS-COCO  \\
\midrule
IntegrAI (ours) & (0.02,\textbf{0.24}) & \textbf{(0.17, 0.53)} & (\textbf{0.07},\textbf{0.43})  \\
DOMINO \cite{eyuboglu2021domino}& (\textbf{0.05},0.20) & (0.11,0.45) & (0.04,0.35)  \\
K-MEANS \cite{rajani2022seal} & (\textbf{0.05},0.20) & (0.09,0.35) & (0.03,0.27) \\
DoubleGreedy \cite{mozannar2022teaching} &(0.01,0.21) &(0.04,0.36) &  (0.03,0.29)  \\
\bottomrule
\end{tabular} 
  \end{adjustbox}

  \end{minipage}
\end{table}

\paragraph{Learning Accurate Integrators ( Aim 1).} The goal is to measure the ability of our method in learning integration functions that lead to low Human-AI team error (the loss $L(H,M)$). This can be well represented by measuring the errors on the training set (discovering regions of error) and the test set (generalization ability). In Table \ref{tab:train_error_real}, we show the results of our method and the baselines at learning integrators and find that our method can find regions that are more informative with respect to Human vs AI performance on the test data.
Figure \ref{fig:bdd_real_plot} shows that on BDD our method can find an integrator that leads to lower loss at test time than the baselines with a minimal number of regions. 

\paragraph{Recovering Ground truth Regions (Aim 2).} We just established that the regions discovered by our algorithm result in a Human-AI team with lower error than human or AI alone. However, it still needs to be verified if the regions are indeed meaningful and consistent regions of space. We utilize a synthetic setup by simulating the AI model and the human responses such that there exist (randomized) regions in the data space where either the human or the AI are accurate/inaccurate (though the regions may slightly overlap). These regions are defined in terms of metadata. As an example on the BDD dataset, we can define the AI to be good at daytime images and bad at images of highways, and the human to be good at nighttime images and bad at images of city streets. We employ our algorithm and the baselines to discover regions and compare them with the ground truth regions corresponding to the partition of the data, which is essentially a clustering task with ground truth clusters. Results are shown in Table \ref{tab:clust_synth} and show that we have clustering metrics mostly higher than the baselines.

\paragraph{Describing Regions (Aim 3).} We conduct an ablation study where we evaluate the power of the contrasting and self-correcting ability of Algorithm~\ref{alg:describe} against baselines. On the MS-COCO dataset, we take regions defined in terms of the presence of a single object (e.g., `apple') and try to obtain a single-word description of the region from the image captions. We use standard captioning metrics that compare descriptions from the algorithms to the object name, we include a metric called "sent-sim" that simply measures cosine similarity with respect to a sentence transformer \cite{reimers-2019-sentence-bert}. We compare to ablations of Algorithm \ref{alg:describe} with $m \in \{0,5,10\}$ (rounds of iteration) and without having examples outside the region (\algname, $S^-=\emptyset$). Results are in Table \ref{tab:captioning} and show that including examples outside the region improves all metrics while increasing iterations ($m$) further improves results slightly.  For the apple example, \algname ($S^-=\emptyset$) finds the description to be ``fruit'' whereas our \algname ($m=5$) finds it to be ``apple''.

\begin{table}[h]
\centering
\caption{Evaluation of our region description algorithm (Algorithm \ref{alg:describe}) on selected subsets of MS-COCO where the different algorithms try to describe a set of images that all contain a given object. For example, a region may be defined by images containing the object ``apple''. Then we compare the descriptions resulting from the different algorithms to the description ``apple''.}
\label{tab:captioning}
    \resizebox{1.0\textwidth}{!}{
\begin{tabular}{l|cc|cccc}
\toprule
& best-caption & SEAL & \algname ($S^-=\emptyset$) & \algname (m=0) & \textbf{\algname (m=5)} & \algname (m=10) \\
\midrule
METEOR & $12.9 \pm 1.9$ & $9.16 \pm 1.89$ & $24.3 \pm 3.3$ & $25.4 \pm 3.2$ & $\mathbf{26.1 \pm 3.3}$ & $25.4 \pm 3.3 $ \\
sent-sim & $39.8 \pm 1.9$ & $44.1 \pm 2.5$ & $65.1 \pm 3.2$ & $67.0 \pm 3.1$ & $66.0 \pm 3.2$ & $\mathbf{68.0 \pm 3.3}$ \\
ROUGE & $5.81 \pm 1.2$ & $0.0 \pm 0.0$ & $25.6 \pm 4.9$ & $32.6 \pm 5.4$ & $27.9 \pm 5.1$ & $\mathbf{35.6 \pm 5.5}$ \\
SPICE & $12.7 \pm 1.9$ & $7.53 \pm 2.3$ & $41.1 \pm 5.8$ & $43.8 \pm 5.8$ & $\mathbf{45.2 \pm 5.8}$ & $\mathbf{45.2 \pm 5.8}$ \\
\bottomrule
\end{tabular}
}
\end{table}

\section{User Studies to Evaluate Onboarding Effect}\label{sec:studies}

\paragraph{Tasks.} We perform user studies on two tasks: 1)   predicting the presence of a traffic light in road images from the BDD dataset \cite{xu2017end} and 2) answering multiple-choice questions from the MMLU \cite{hendrycks2020measuring} dataset. For BDD, we blur the images with Gaussian blur to make them difficult for humans and use the Faster R-CNN model as the AI. Participants can see the AI's prediction, bounding box on the image as an explanation and the model's confidence score. For MMLU, participants are shown a question, four possible answers and the prediction of GPT-3.5-turbo \cite{openai2020chatgpt}, and then have to pick the best answer. We also obtain an explanation from GPT-3.5 by using the prompt ``Please explain your answer in one sentence.'' Both the AI answer and the explanation are shown. 
GPT-3.5 obtains an accuracy of 69\% during our evaluation and we restrict our attention to specific subjects within the MMLU dataset. Specifically, we sample 5 subjects (out of the 57 in MMLU) where ChatGPT has significantly better performance than average, 5 where it's significantly worse, and 4 subjects where performance is similar to average performance. We sample 150 questions from each subject and additionally sample 150 questions from the OpenBookQA dataset \cite{mihaylov2018can} to use as attention checks. We show the prediction interfaces in Figure \ref{fig:interface}. Details are in Appendix \ref{apx:user_studies}.

\begin{figure}[t]
    \centering
    \begin{subfigure}{0.48\textwidth}
        \centering
          \includegraphics[width=\textwidth]{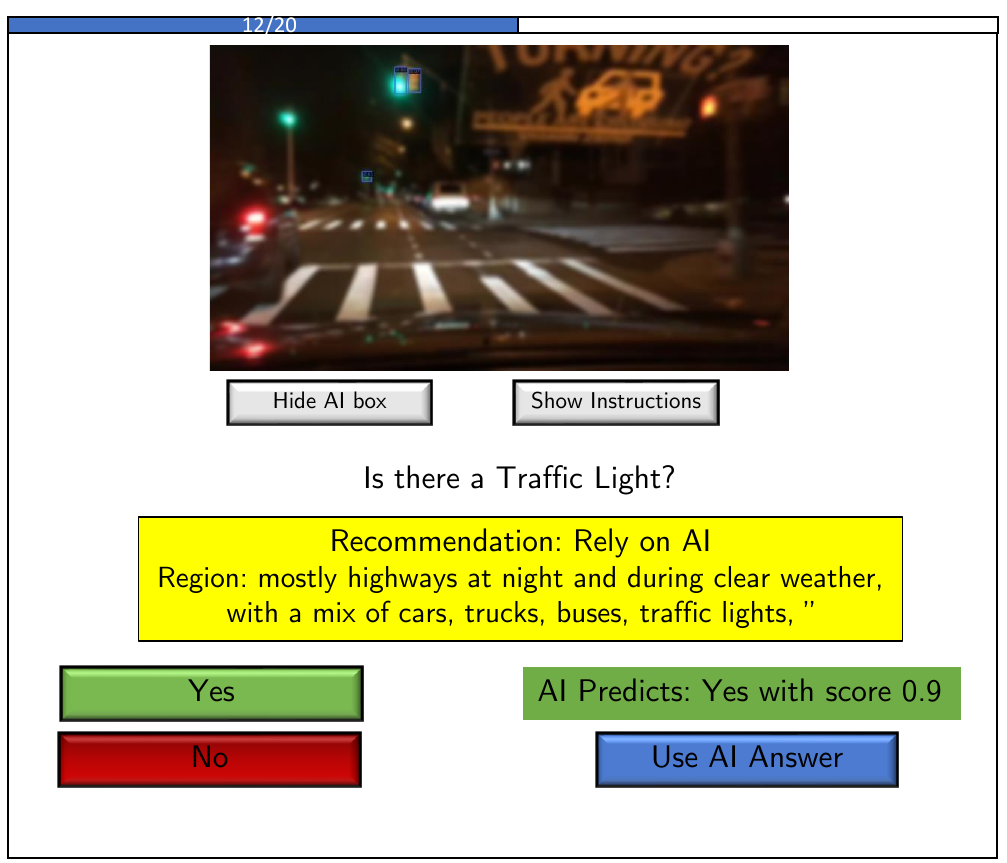}
        \subcaption{BDD}
        \label{fig:bddui}%
    \end{subfigure}\hfill%
    \begin{subfigure}{0.48\textwidth}
        \centering
          \includegraphics[width=\textwidth]{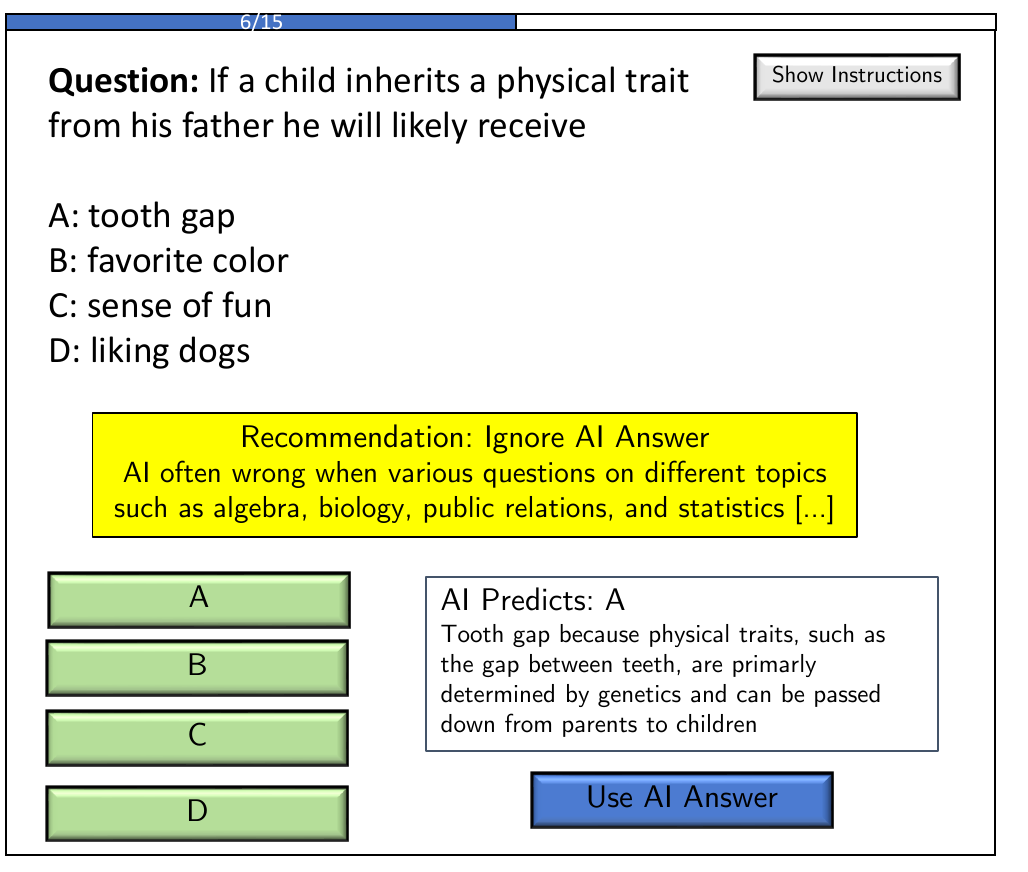}
        \subcaption{MMLU}
        \label{fig:mmluii}%
    \end{subfigure}
    \caption{(a) Interface for humans to detect a traffic light in images from BDD dataset in the presence of AI's prediction, confidence score, and bounding box and (b) interface for humans to answer multiple choice questions from MMLU dataset with AI's prediction and explanation.}
      \label{fig:interface}
      \vspace{-0em}
\end{figure}

\paragraph{Participants.} We submitted an IRB application and the IRB declared it exempt as is. All participants agreed to a consent form for sharing study data. We recruited participants from the crowdsourcing website Prolific \cite{prolific} from an international pool, filtering for those who are fluent in English, have above a 98\% approval rating, have more than 60 previous submissions, and have not completed any of our studies before. For BDD, participants are compensated  \$3 per 20 examples in the study and then some receive a bonus of \$2 for good performance. For MMLU, we pay participants \$3 for every 15 questions. We collected information about participants' age, gender (52\% identify as Female), knowledge of AI, and other task-specific questions. Participants have to correctly answer on three initial images without blur in the case of BDD (questions from OpenBookQA for MMLU). They encounter attention checks throughout the study to further filter them as we exclude participants who fail all attention checks. 

\paragraph{Experimental Conditions.} For BDD, we initially collect responses from 25 participants who predict without the AI and then predict with the help of AI (but no onboarding). We use this data as the basis of the dataset $D$ of prior human integration decisions and human predictions to find 10 regions using \algname.  We then run four different experimental conditions with 50 unique participants in each where each participant predicts on 20 examples: (1) human predicts alone (H) and human predicts with the help of AI (H-AI)  but no onboarding and random order between with and without AI, (2) human receives onboarding using our method and then in a random order also receives recommendations (Onboard(ours)+Rec) or no recommendations (Onboard(ours)), (3) human goes through a modified onboarding procedure that only uses step 1 and step 2 from Section \ref{sec:onboarding} and then uses regions from DOMINO \cite{eyuboglu2021domino} (Onboard(baseline)), and finally (4) human does not receive onboarding but receives the AI-integration recommendations (Rec). For MMLU, participants are tested on 15 examples per condition and onboarding goes through 7 regions found by our algorithm. We run \algname twice: once on both the dataset embeddings and metadata (subject name), and once on an embedding of ChatGPT explanations separately. We find 10 regions based on the metadata and 2 regions based on the ChatGPT explanations, we show participants the 7 regions that have the highest gain. Due to budget constraints, we only run conditions 1-2-4 for MMLU. 
Note that all participants receive the introduction phase of the onboarding (AI model information) regardless of whether they go through the teaching phase or not. 

\paragraph{Results.} In Table \ref{fig:bdd_study_results} and Table \ref{fig:mmlu_study_results} we display various results from the user studies for BDD and MMLU respectively across all experimental conditions. We show the average accuracy ($\pm$ standard error) across all participants of their final predictions, AI reliance as measured by how often they pressed the ``Use AI Answer'' button, and the average time it took for them to make a prediction per example (we remove any time period of more than 2 minutes). Finally, we compute using a two-sample independent t-test the p-value and t-test statistic when comparing each condition (the columns) to the Human-AI condition where the human receives no onboarding (to compare the Human-only condition to Human-AI we use a paired t-test for this pair only). Since we perform multiple tests, we need to correct for multiple hypothesis testing so we rely on the Benjamini/Hochberg method \cite{benjamini1995controlling}.

\paragraph{Analysis.} For BDD, we first observe that human and AI performance are very comparable at around 79\%, which reduces slightly to 77.2\% when the human collaborates with the AI without onboarding. Participants who go through onboarding have a significantly higher task accuracy compared to those who didn't go through onboarding (corrected p-value of 0.042) with a 5.4\% increase. The onboarding baseline fails to significantly increase task accuracy, showcasing that the increase is not just due to task familiarity but possibly due to insights gained from regions found by \algname. Displaying recommendations in addition to onboarding (Onboard(ours)+Rec) does not improve performance but adds time to the decision-making process (7.6s compared to 5.9s without). For MMLU, we note that there is a 20\% gap between human and AI performance, but human+AI with and without onboarding can obtain an accuracy of around 75\% which is slightly higher than AI alone; onboarding had no additional effect. Onboard(ours) does result in slightly lower time per example than Human+AI without onboarding. Interestingly, we find a weakly significant negative effect of only showing AI-integration recommendations, which decreases accuracy by 5\% and adds 6 seconds of time per example.

\begin{table}
\centering
\resizebox{\textwidth}{!}{
\begin{tabular}{l|lll|llll}
\toprule
Metric & AI only & Human & Human+AI & Onboard(ours)+Rec & Onboard(ours) & Onboard(baseline) & Rec \\
\midrule
Accuracy (\%) & $79.0 \pm 0.7$ & $78.5 \pm 1.7$ & $77.2 \pm 1.4$ & $79.9 \pm 1.4$ & $82.6 \pm 1.3$ & $80.4 \pm 1.4$ & $81.4 \pm 1.8$ \\
Test vs H-AI & $0.272, 1.211$ & $0.455, 0.752$ & N/A & $0.268, 1.352$ & $0.042, 2.747$ & $0.261, 1.525$ & $0.206, 1.839$ \\
AI reliance (\%) & N/A & N/A & $16.5 \pm 3.1$ & $66.5 \pm 2.4$ & $25.5 \pm 3.4$ & $24.4 \pm 4.4$ & $21.4 \pm 3.2$ \\
Time/example (s) & N/A  & $5.408 \pm 0.289$ & $7.78 \pm 0.517$ & $7.622 \pm 0.371$ & $5.936 \pm 0.288$ & $6.841 \pm 0.543$ & $8.717 \pm 0.516$ \\
\bottomrule
\end{tabular}}
\caption{Results from our user studies for BDD. For accuracy, time per example, and AI-reliance we report mean and standard error across participants. The "Test vs H-AI" row reports the adjusted p-value and t-test statistic for a two-sample t-test between the human+AI condition and the other conditions (columns). }
\label{fig:bdd_study_results}
\end{table}

\begin{table}
\centering
\resizebox{\textwidth}{!}{
\begin{tabular}{l|lll|lll}
\toprule
Metric & AI only & Human & Human+AI & Onboard(ours)+Rec& Onboard(ours) & Rec  \\
\midrule
Accuracy (\%) & $72.9 \pm 0.6$ & $52.8 \pm 2.2$ & $75.0 \pm 1.7$ & $73.7 \pm 1.8$ & $74.4 \pm 1.7$ & $69.8 \pm 1.8$  \\
Test vs H-AI & $0.230, -1.488$ & $0.0, -7.899$ & N/A & $0.747, -0.53$ & $0.792, -0.265$ & $0.101, -2.08$ \\
AI reliance (\%) & N/A & N/A & $40.0 \pm 3.6$ & $34.5 \pm 3.7$ & $40.6 \pm 3.8$ & $34.2 \pm 2.9$ \\
Time/example (s) & N/A  & $30.608 \pm 2.109$ & $23.623 \pm 1.66$ & $22.917 \pm 1.362$ & $20.977 \pm 1.509$ & $29.535 \pm 1.883$ \\
\bottomrule
\end{tabular}
}
\caption{Results from our user studies for MMLU. }
\label{fig:mmlu_study_results}
\end{table}

\paragraph{(Informal) Qualitative Analysis.} At the end of each experiment, we asked the participant the following question: ``What was your decision process for relying on the AI answer and for when to ignore the AI answer?'' For the BDD task, we compare the responses of participants who were in the baseline Human+AI condition versus participants in the Onboard(ours) condition. To summarize the responses, we use \algname-Describe with $m=0$,\footnote{No iteration because we can fit all responses in the context of the LLM which is GPT-4 in this analysis. We changed the prompt to ask for 100 words instead of the usual 20-word limit.} first with the region of interest corresponding to the responses from the Onboard(ours) condition (asking it to contrast with Human+AI responses). We get the following description verbatim (and add bolding for emphasis):
\begin{quote}
Points inside the region involve scenarios primarily where the individual uses or relies on AI when uncertain, particularly when visibility is poor or objects are too distant. In contrast, points lying outside the region pertain to cases where \textbf{individuals often ignore the AI} either because they feel confident in their own judgment, the picture is clear, or they believe the AI is not accurate enough.
\end{quote}
On the other hand, when we describe the region corresponding to Human+AI responses, we get:
\begin{quote}
The region comprises descriptions wherein individuals primarily rely on their own judgement to identify traffic lights in images, resorting to the AI's aid when the image is too blurry, unclear or when doubt exists. In stark contrast, points outside the region detail instances where reliance on AI is more pronounced or where\textbf{ external factors like road type and presence of cars are considered.}
\end{quote}
One theme that emerges is using AI more in the onboarding condition, which is confirmed quantitatively as participants in the Onboard(ours) condition relied on AI 9\%  more (see Table \ref{fig:bdd_study_results}). Another theme is that of participants in onboarding relying on external factors, which can potentially be attributed to lessons learned during onboarding.

\paragraph{Discussion.} We believe that for MMLU, due to the wide gap between human and AI accuracy and the availability of GPT-3.5 explanations, onboarding did not improve performance. We note that in a lot of instances, GPT-3.5 explanations express uncertainty over the answer, as in the following examples:
\begin{quote}
   - `Unfortunately, the options provided do not provide a clear answer to what happens after the Meyer tree's flower petals drop. Can you please provide more information or context about the question'
   
   - `The answer cannot be provided with the given information as it does not specify which president is being referred to.'
   
   - `[...] the answer cannot be provided with the given information as it does not specify which president is being referred to.'
\end{quote}
Such statements about uncertainty can already help the human more accurately know when not to trust the AI and try harder to find the correct answer. In cases where GPT-3.5 does not express uncertainty, it tries to explain its answer and often does so correctly, but it is not clear whether the explanations allow the human to easily verify the answer.
Moreover, it is clear that in its current form, displaying the AI-integration recommendation is not an effective strategy and that onboarding on its own is sufficient. Finally, note that even the Human-AI baseline benefits from the human-AI card, which might explain why the team is at least as good as its components in our experiments. 

\paragraph{Limitations.} Onboarding and recommendations can significantly affect human decision making. If the recommendations are inaccurate, they could lead to drops in performance and thus require safeguarding. Onboarding and recommendations can be tailored to the specific human by leveraging their characteristics to few-shot learn their prior integrator and prediction abilities. 

\section{Conclusion}

In this paper, we introduced \algname, a novel algorithmic framework designed to enhance human-AI collaboration. At the core of our framework is the AI-integration function which represents the human's mental model to either rely on, ignore, or collaborate with AI on a task-specific basis. The first step of our algorithm is to collect data about human performance on the task and about their prior expectations and reliance on the AI model. The objective is to teach the human in order to correct their prior about how to cooperate with the AI. The second step is to discover regions of the data as local neighborhoods in an embedding space where the human prior is incorrect. The third step is to describe the regions with natural language and label them with the correct action the human should take in that region: either use or ignore the AI. Finally, in the onboarding phase, we teach these regions to the participants using our proposed method. We found that on an object detection task that our onboarding procedure significantly improved performance and that in another question-answering task it did not have a significant effect. 

There are many directions for future work to explore both on the algorithmic side and on the behavioral side. For instance, our region discovery and region description algorithms are decoupled, ideally we can jointly discover and describe regions so that only regions with interpretable language descriptions are found. Second, our user studies only handled the case when the human can either use or ignore the AI, collaboration with the AI was implicitly done by the participants but it was not captured in our data, future work can build interfaces and procedures to handle the case when $R=2$. In our user studies, we only allowed participants to discuss how they used the AI  after the study was completed, we did not collect enough data about user behavior and user learning, future work could explore more qualitative insights about onboarding.

\section*{Acknowledgments}
We thank Hunter Lang and Arvind Satyanarayan for their feedback on the early stages of this work. HM is thankful for the support of the MIT-IBM Watson AI Lab.
\bibliographystyle{alpha}
\bibliography{ref}
\appendix
\addcontentsline{toc}{section}{Appendix} 
\part{Appendix} 
\parttoc 
\section{Extended Related Work}\label{apx:extend_related_work}

Reference~\cite{lai2022human} proposes a method for human-AI collaboration via conditional delegation rules that the human can write down. Our framework enables the automated learning of such conditional delegation rules for more general forms of data that can also depend on the AI output. \cite{vodrahalli2022uncalibrated} proposes to modify the confidence displayed by the AI model to appropriately encourage and discourage reliance on the AI model. However, this technique deliberately misleads the human on the AI model ability, our methodology incorporates similar ideas by learning the human prior function of reliance on the AI and then improving on it with the learned integration recommendations, however, we display these recommendations in a separate dashboard without modifying the AI model output. 
A related approach to our methodology by \cite{ma2023should} is to adaptively display or hide the AI model prediction and display the estimated confidence level of the human and the AI on a task of predicting whether a person's income exceeds a certain level. They show that displaying the confidence of the human and the AI to the human improves performance. Our method is able to learn the confidence level of the human and the AI, but also incorporates how the human utilizes the AI and describes the regions where AI vs human performance is different. \cite{cabrera2023improving} presents a similar approach to our AI recommendations, however, they use simulated and faked AI models and descriptions of behavior while we are able to obtain automated generation of these descriptions of AI behavior.

Existing research has examined various methods to establish human trust in and replicate the predictions of machine learning models. One such method is LIME, a black-box feature importance technique, which was employed to select examples for evaluation by crowdworkers to determine the superior model among two options \cite{ribeiro2016should,lai2020chicago}. However, their selection strategy disregards the human predictor, and their approach merely presents the examples without further action.
In the context of visual question answering, Chandrasekaran et al. \cite{chandrasekaran2018explanations} manually selected seven examples to educate crowdworkers about the AI's capabilities, leading to an enhanced ability to identify instances where the AI failed. Feng et al. \cite{feng2019can}, in the domain of Quizbowl question answering, emphasize the significance of incorporating the human expert's skill level when designing explanations. This further justifies our decision to involve the human predictor in the selection of teaching examples.
Cai et al. \cite{cai2019hello} conducted a study involving 21 pathologists to gather guidelines on what clinicians desired to know about an AI system prior to interacting with it. Yin et al. \cite{yin2019understanding} investigated the impact of initial debriefing on stated AI accuracy versus observed AI accuracy during deployment, finding a substantial influence of stated accuracy on trust that diminishes quickly once the model is observed in practical use. This reinforces our approach of building trust through examples that simulate real-world deployment.
Bansal et al. \cite{bansal2019beyond} examined the role of the human's mental model of the AI in task accuracy; however, the mental model was developed through interaction during testing rather than during an initial onboarding stage.
The most similar work to ours is that of \cite{mozannar2022teaching} which presents an onboarding scheme based on selecting a set of examples and allows the human to describe the regions where the AI performance is good or bad. Through a user study on passage-based question answering, they show that their onboarding scheme improves performance by 5\%, however, they evaluate without the presence of AI evaluations, with a synthetic AI model and their scheme involves more involvement from the human as they have to describe the regions themselves. Another approach to teaching involves providing humans with guidelines on when to rely on AI systems \cite{amershi2019guidelines}. Model cards \cite{mitchell2019model} and industry practices such as the IBM AI fact sheet \cite{arnold2019factsheets} demonstrate direct methods of presenting these guidelines to users, we present humans with a similar form of card but that includes aspects of human performance "human-AI card".

There is a growing and  large area of literature on discovering (and auditing)  regions of AI error, the following is not meant as an extensive list of related work but captures some of the essence of the literature:
\begin{itemize}
\item Adatest allows a user to iteratively discover regions of AI error using  LLMs  for NLP tasks and then re-train the model on the regions of error  \cite{ribeiro2022adaptive}, it was then extended to vision tasks \cite{gao2022adaptive} with a similar procedure in \cite{wiles2022discovering}.
\item Erudite allows users to discover regions of error of NLP models through user interfaces \cite{wu2019errudite}, there is a wider literature on dashboards for discovering regions of error \cite{ahn2023escape}
\item  \cite{jain2022distilling} learns an SVM model from image embeddings to predict model error and then uncover regions of error based on the directions of the SVM model.

 \item Works have done extensive manual annotation of ImageNet model mistakes\cite{idrissi2022imagenet,vasudevan2022does}. 
 \item DOMINO discovers regions of model error using a slice discovery model based on a specialized gaussian mixture model \cite{eyuboglu2021domino}, extensions include DRML \cite{zhang2022drml}.
    \item The Spotlight method learns individual regions based on a neighborhood of a learned point \cite{d2022spotlight}, our region finding algorithms generalize their procedure by learning weighted distance distance measures and with a different aim of improving gain of the prior.  
    \item SEAL: Interactive Tool for Systematic Error Analysis and Labeling, uses k-means to uncover regions of error and then uses an LLM to describe each region \cite{rajani2022seal}.
\end{itemize}

Recent work has emerged on describing sets of images  \cite{hupert2022describing,malakan2022vision} but they don't incorporate a contrastive method as we propose. Helpful tools for describing differences between text and images can be useful for describing regions which future work can incorporate \cite{zhong2022describing,wang2020neighbours,wang2022distinctive}.

One of the objectives of explainable machine learning is to enhance humans' ability to assess the accuracy of AI predictions by offering supporting evidence \cite{lai2019human,lage2019evaluation,smith2020no,hase2020evaluating,zhang2020effect,kocielnik2019will,suresh2020misplaced,suresh2021intuitively,wortmanvaughan2021a,gonzalez2020human}. Nevertheless, these explanations fail to provide guidance to decision makers on how to balance their own predictions against those of the AI or how to integrate the AI's evidence into their final decision \cite{kaur2020interpreting}. \cite{vasconcelos2023explanations} shows that AI explanations can reduce overreliance and improve human-AI team performance, however, their experiments are with simulated AI models and explanations. The central question of our work is not to study the utility of AI explanations, in fact, all our user studies incorporate AI explanations and we aim to improve human-AI performance in their presence.


\section{Region Finding Algorithm - Details}\label{apx:region_finding}

\paragraph{Regions Requirements.} Each region in our algorithm should aim to satisfy the following constraints:

\begin{enumerate}
	\item \textbf{Region Size:} We want the size of the region to be at least of size $\beta_l$ and at most of size $\beta_u$.
	\item \textbf{Consistency of takeaway:} 
	The examples in each region must agree on what the takeaway is in terms of the integration decision. Specifically, at least $\alpha$\% of all points in the region must either be: ignore AI, use AI as is or integrate AI advice.
	\item \textbf{Concise and Distinguishable Theme:} Each region must be concisely described in natural language in such a way to differentiate from the overall domain. If a region cannot be described in natural language, the human may not be able to derive a generalizable recommendation from it. This is a constraint that we implicitly try to satisfy by learning neighborhoods in a natural language embedding space.
	\item \textbf{Minimum Gain:} Each region must have a minimum information gain (defined below) of $\delta$. This is to ensure that all regions contain sufficient novel information to the human.
\end{enumerate}

The optimization to find each region can be formulated in its non-relaxed form as:
\begin{align*}
    & \max_{c, \gamma, w, r} \ \ \sum_{i=1}^n \bI_{ ||w((e_i,a_i)-c)|| < \gamma}  \cdot \bm{g}_{i,r}  \\ 
    & \textrm{s.t.}  \quad  \sum_{i=1}^n \bI_{ ||w((e_i,a_i)-c)|| < \gamma}  \cdot   \bI_{r_i = r} \geq \alpha n \\
    & \textrm{s.t.} \quad  n \beta_l \leq \sum_{i=1}^n \bI_{ ||w((e_i,a_i)-c)|| < \gamma}  \leq  n \beta_u  
\end{align*}

And the relaxation we propose is (refer back to Section \ref{sec:learn_rules:discovery}):

\begin{align*}
  &\resizebox{\linewidth}{!}{$ \max_{c, \gamma, w, r} \ \ \sum_{i=1}^n \sigma(C_1(-||w \circ ((e_i,a_i)-c)|| + \gamma))  \cdot g_{i,r}  - \lambda \max ( \sum_{i=1}^n \sigma(C_1(-||w \circ ((e_i,a_i)-c)|| + \gamma))   $}\\
    &\cdot ( \bI_{r^*_i = r}  - \alpha n)  ,0)- \lambda \max\left( \sum_{i=1}^n \sigma(C_1(-||w \circ ((e_i,a_i)-c)|| + \gamma))  - \beta_u n ,0\right) \\
    &  \lambda \max\left( -\sum_{i=1}^n \sigma(C_1(-||w \circ ((e_i,a_i)-c)|| + \gamma))  + \beta_l n ,0\right)
\end{align*}

We run the optimization for $r=0$ and $r=1$ and choose the $r$ with the better objective value. With $r$ fixed, we optimize with respect to the remaining continuous parameters using AdamW and reduce the learning rate when loss has stopped improving. We initialize with $\gamma =0$ and $w = \bm{1}$. For the centroid $c$, we first run $k$-medoids clustering on the input data with $k=\min(\max(100, T), n)$ and randomly select 20 of the resulting centroids. Then with $c$ initialized as each one of the 20 centroids in turn, we run the optimization for 200 epochs and record the loss, and finally optimize for 2000 epochs with the best initialization for $c$. We do these repeated initializations to avoid local minima which are a common failure mode of this type of optimization problem. Note this process is to find one region, to find all regions we repeat this process identically to find regions one by one.

\paragraph{Selection Based Approach.} We described in the body of the paper a generative algorithm to find the regions. We now describe a selection based algorithm that finds centroids from points in the dataset $\tilde{D} = \{e_i,r_i\}$. We will also restrict the radius to be the distance between the centroid to another data point in $\Tilde{D}$. We proceed with a sequential search, at round $i$ we perform the following search:
\begin{align}
	c_i, \gamma_i   =& \arg \max_{i \in \tilde{D}, \gamma }  G( N_{i,\gamma} , \hat{R}^*, R_{N_{1:i-1}}) ,    \\
	& \text{s.t.} \ \exists k \in [n] \ s.t. \ \gamma = d(e_i, e_k), \label{eq:gamma_in_data} \\
	& \text{and} \  \frac{\sum_{j \in [n], d(e_i,e_j) < \gamma} \bI_{r_j = r_i}}{ |\{j \in [n], d(e_i,e_j) < \gamma\}|} \geq \alpha  \label{eq:alpha_constraint} \\
	&  \text{and} \ \beta_{l} \leq  |\{j \in [n], d(e_i,e_j) < \gamma \}| \leq \beta_{u} \\
    &  \text{and} \  G( N_{i,\gamma} , \hat{R}^*, R_{N_{1:i-1}}) > \delta
\end{align}
Note that with the selection-based procedure, we have to define a fixed distance measure $d$, and we cannot optimize over the deferral decision $r$ of the region as we inherit from the point $i$ found. The naive algorithm to solve the above search is as follows. We first compute the distance matrix between all data points in $\Tilde{D}$, call this matrix $K$. We then at each round do the following for each point $i$ in $D\Tilde{D}$: sort the points by their distance to $i$, iteratively grow the region around the point $i$ to satisfy the constraints, and then keep track of the maximum gain radius. Each time we grow the region by one point, we check if the constraints are satisfied. The search is parallelized across multiple instances to make it faster. Finally, we compare the gain of all feasible points and pick the highest one. This algorithm extends the approach of \cite{mozannar2022teaching} to incorporate additional constraints and has significant speed ups over their approach.

\paragraph{Aggregating Regions Across Different Embedding Spaces}. We can run our region finding algorithm above and find different regions across multiple embedding spaces. The question is how do we aggregate these regions. Say we found a set of $T$ regions $N_1,\cdots,N_T$. We then run the following meta-algorithm: do a greedy sequential search to add regions one at a time while making sure the minimum gain requirement is satisfied, stop when it is not. 

\section{Region Description Algorithm - Details}\label{apx:region_describe}

The LLM call inside Algorithm 1 is accomplished by building the prompt with $S^+$ being the inside set (positives) and $S^-$ being the outside set (negatives) as follows: 

\begin{python}
def get_prompt(positives, negatives):
    prompt = pre_instruction + "\n"
    prompt += "inside the region: \n "
    counter = 1
    for p in positives:
        prompt += p[0] + ", \n "
        counter += 1
    if len(negatives) > 0 :
        prompt += ". \n not in the region: \n"
        counter = 1
        for p in negatives:
            prompt += p[0] + ",\n"
            counter += 1
    prompt += post_instruction
    return prompt
\end{python}

For the experiments in Section \ref{sec:method_eval} we use the following pre instruction:
\begin{quote}
I will provide you with a set of descriptions of points that belong to a region and a set of descriptions of point that do not belong to the region. Your task is to summarize the points inside the region in a concise and precise short sentence while making sure the summary contrasts to points outside the region. Your one sentence summary should be able to allow a person to distinguish between points inside and outside the region while describing the region well. The summary should not be a single word, it should be accurate, concise, distinguishing, and precise.

Example:

Inside the region:

- two cows and two sheep grazing in a pasture.

- the sheep is standing near a tree.

Not in the region:

- the cows are lying on the grass beside the water.

summary:
sheep.

End of Example

\end{quote}
While the post-instruction is simply "summary:". 

For the ablation without contrasting, the pre instruction we use is:
\begin{quote}
    I will provide you with a set of descriptions of points that belong to a region.Your task is to summarize the points inside the region in a concise and precise short sentence .Your one sentence summary should be able to allow a person to distinguish  points inside the region while describing the region well.The summary should be a single word, it should be accurate, concise, distinguishing and precise.
    
    Example: 
    
    inside the region:
- two cows and two sheep grazing in a pasture.

-the sheep is standing near a tree.

summary: sheep.

End of Example 

\end{quote}

For the user studies, we had used an earlier instruction that works slightly worse and post-processed the descriptions by only modifying the first few words to be consistent (see the real examples in later sections):
\begin{quote}
"summarize the points inside the region in a concise and precise short sentence while making sure the summary contrasts to points outside the region"
\end{quote}

We recommend the use of the following pre instruction:
\begin{quote}
    I will provide you with a set of descriptions of points that belong to a region and a set of descriptions of points that do not belong to the region. Your task is to summarize the points inside the region in a concise and precise short sentence while making sure the summary contrasts to points outside the region. Your one sentence summary should be able to allow a person to distinguish between points inside and outside the region while describing the region well. The summary should be no more than 20 words, it should be accurate, concise, distinguishing and precise.
    
    Example: 
    
    inside the region: 
    
    - two cows and two sheep grazing in a pasture. 
    
    - the sheep is standing near a tree. 
    
    outside the region: 
    
    - the cows are lying on the grass beside the water.
    
    summary: The region consists of descriptions that have sheep in them outside in nature, it could have cows but must have sheep. 
    
    End of Example 

\end{quote}

\section{Onboarding and Recommendations to Promote Rules - Details}\label{apx:onboarding}

\paragraph{Introductory Phase.} To facilitate a smooth onboarding process for individuals working with an AI assistant, we introduce the Human-AI Card. This card provides detailed insights into the AI's capabilities, training, and performance. 

Additionally, we provide a breakdown of AI and Human performance on different subgroups of data.
We take an example of the Berkeley Deep Driving dataset, where a subgroup might comprise images taken during the night, in rainy weather, or on a highway. We compute the model's error for each possible subgroup and then perform a paired t-test comparing the subgroup model error to the average model error over the entire data. For the purpose of our user studies, we highlight subgroups defined by a single metadata category that show statistically significant differences ($p \leq 0.05$). It's important to note that, for rigorous analysis, one should apply corrections for multiple hypothesis testing. However, considering the vast number of metadata categories, many results might become insignificant. Therefore, for simplicity, we adopt this heuristic approach.

\section{Method Evaluation - Details}\label{apx:method_eval}
In Table \ref{tab:datasets} we report the details on the datasets we use in our method evaluation.  We normalize all datasets using $l_{\infty}$ normalization and run our algorithms for 2000 epochs with a learning rate of 0.001 using AdamW and a scheduler to update. We use a constant $C_1=20$.

\begin{table}[H]
    \centering
        \caption{Datasets for "Learning Accurate Integrators ( Aim 1)". We note the total number of samples $n$, the target set size $|\mathcal{Y}|$, the human expert finally the model of the AI. When we note human "XX\% accurate", this indicates a synthetic human model that is accurate uniformly at random with probability XX\%. For DynaSent, the AI model is a a pre-trained sentiment analysis roBERTa-base model \cite{barbieri2020tweeteval} on Twitter data and achieves 75\% accuracy. For both BDD and MS-COCO we blur the images using a Guassian Blur with scale 21 and variance 5. }
    \resizebox{1\textwidth}{!}{
    \begin{tabular}{p{0.2 \textwidth}c{p}{0.1 \textwidth}lp{0.2 \textwidth}}
    \toprule
         \textbf{Dataset} & $n$ & $\left| \mathcal{Y} \right|$ &\textbf{Human} & \textbf{AI} \\ 
         \toprule
Berkeley Deep Drive (BDD) \cite{xu2017end,bdd100k}  &10k &2 & 80\% accurate & faster rcnn r50 fpn 1x \footnote{\url{https://github.com/SysCV/bdd100k-models}} -  Gaussian blur with scale 21 and variance 5\\
MS-COCO \cite{lin2014microsoft}  &5k &2 (presence of person in image) & 70\% accurate &  faster rcnn R 50 FPN   \\
Massive Multi-task Language Understanding (MMLU) \cite{hendrycks2020measuring}  &14k &4 (MCQ) &50\% accurate & flan-t5xl \cite{chung2022scaling}\\
Dynamic Sentiment Analysis Dataset (DynaSent) \cite{potts2020dynasent}  &6.5k &3 & leave-one-out annotator& a pre-trained sentiment analysis roBERTa-base model \cite{barbieri2020tweeteval} \\
\\ \bottomrule
    \end{tabular}}
    \label{tab:datasets}
\end{table}

\begin{figure}[H]
    \centering
    \includegraphics[width=0.7\textwidth]{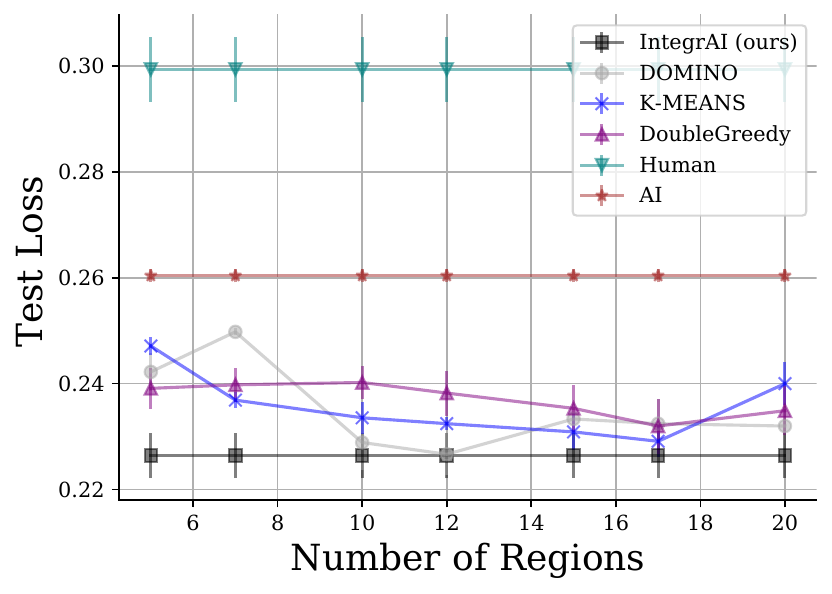}
    
    \caption{Test Error ($\downarrow$) of the human-AI system when following the decisions of the different integrators baselines as we vary the number of regions maximally allowed for each integrator on the MS-COCO dataset.  }
    \label{fig:coco_real_plot}
\end{figure}

\begin{figure}[H]
    \centering
    \includegraphics[width=0.7\textwidth]{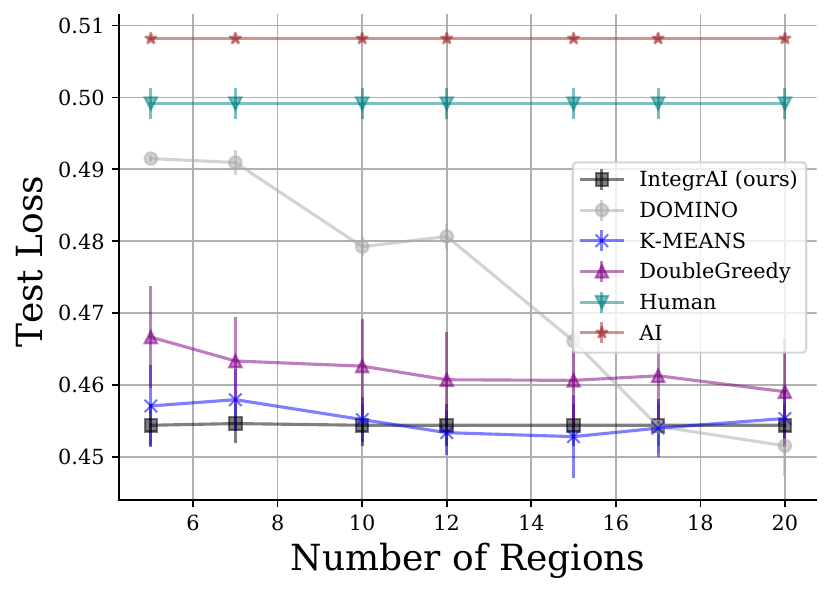}
    
    \caption{Test Error ($\downarrow$) of the human-AI system when following the decisions of the different integrators baselines as we vary the number of regions maximally allowed for each integrator on the MMLU dataset.  }
    \label{fig:mmlu_real_plot}
\end{figure}

\begin{figure}[H]
    \centering
    \includegraphics[width=0.7\textwidth]{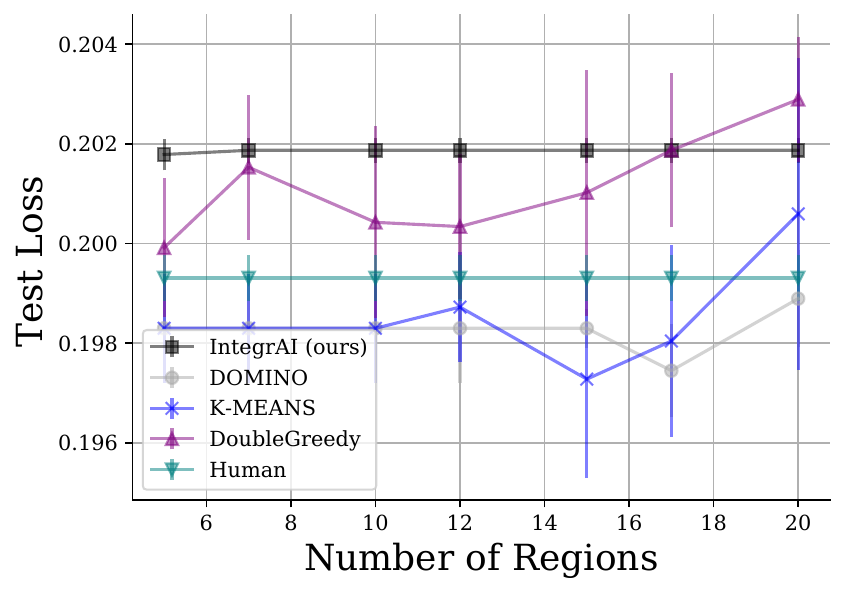}
    
    \caption{Test Error ($\downarrow$) of the human-AI system when following the decisions of the different integrators baselines as we vary the number of regions maximally allowed for each integrator on the Dyanasent dataset.  }
    \label{fig:dynasent_real_plot}
\end{figure}

For Aim 2, we create synthetic AI and human models as follows:
\begin{itemize}
    \item BDD: AI and Human model each have four regions defined as a condition of a randomly selected metadata feature, each region is either a good region where the AI/Human have 95\% accuracy or bad region with 60\% accuracy (equal number of good and bad regions). If an example does not belong to any region the AI/Human have 75\% accuracy. Each region is of size at least $0.01$ and at most $0.2$ in terms of fraction of  data points in the dataset.  An example region is "AI is good at: weather: clear" or "Human is bad at:  timeofday: night ".
    \item MMLU: same setup as BDD, region defined as the subject of example. An example region is "AI is good at: subject: professional psychology".
    \item MS-COCO: same setup as BDD, region defined as presence of object in image. An example region is "'Human is bad at: cow: present".
\end{itemize}

For Aim 3, an obstacle to quantitative results for region descriptions is that they are often complex even when regions are synthetically defined in terms of metadata and captioning metrics are not informative. For a region defined on BDD as images of "scenes of highways during the night with no cars", our algorithm finds the description "highway during the night with various weather conditions and not congested with many cars" while the SEAL \cite{rajani2022seal} describes it as "Highway Nighttime Weather Conditions". The SEAL description surprisingly has higher captioning metric scores (BLEU, METEOR) \cite{luo2022thorough} while being less informative.

For the MS-COCO evaluation, we only select objects that have at least 50 examples present in the evaluation set which leads to only 73 objects over which we evaluate the different region description algorithms.

\section{User Studies - Details}\label{apx:user_studies}

\subsection{BDD Study}

\paragraph{Task.} The images from BDD are blurred using a Gaussian Blur with a scale of 21 and a variance of 5. The AI model is a trained faster rcnn model that achieves 84\% accuracy without blurring which decreases to 78\% accuracy on the test set. We use the bounding boxes from the model and output them on the image (allowing the user to either hide or show them). To get a confidence score, we take the maximum score for the prediction of a traffic light in the image.

\paragraph{Initial Data Collection.} The BDD dataset was split 70-30 where the 70\% split was used to get the initial human predictions and find the regions and the 30\% split was used only to get testing examples for the final user study. As mentioned, we obtained data on 400 examples with both human predictions and prior AI-integration decisions. We use these examples to build Random Forrest models that predict both the human predictions and AI-integration decisions from the embeddings, labels, and AI predictions (AI predictions only for predicting AI integration decisions). We use these predictions from the RF models to label the entire dataset. We ensure that the predictions of the RF models are calibrated (if the human is 80\% accurate, the model is also 80\% accurate) with the human predictions and integration decisions by modifying the threshold on model probability used to make predictions (e.g. from the usual 0.5 threshold to the value that makes the models calibrated). Each participant is evaluated on a randomized set of examples, we create 40 different sets of 20 images that get assigned randomly to each participant. 

\paragraph{Attention Checks.} In each condition, we insert 3  attention checks for every 20 examples where the images are unblurred and we keep the AI prediction. We only retain responses where participants don't get all attention checks wrong. The class balance of the dataset is close to 51\%-49\%. The attention checks are used as part of the study results as they only modify the blur of the images. 

\paragraph{Regions Found.} The regions found by our procedure are shown in Table \ref{tab:regions_bdd}.

\begin{table}[H]
    \centering
        \caption{Region descriptions found by \algname for the BDD user study. }
    \resizebox{1\textwidth}{!}{
    \begin{tabular}{cp{0.8\textwidth}}
    \toprule
Region ID & Description \\
1 &  depict various city streets and highways during the daytime with different weather conditions, containing pedestrians, cars, trucks, traffic lights, and signs., \\ 
2 &   depict various types of roads and streets during the daytime with different weather conditions, containing cars, pedestrians, traffic lights and signs, while the outside examples include scenes with fewer objects or in less common locations such as a parking lot.,\\ 
3 &  depict various scenarios of streets and highways with moderate to heavy traffic flow during the day or night, with different weather conditions, along with traffic signs and lights, cars, trucks, buses, and pedestrians.\\ 
4 &  depict various outdoor scenes during the daytime or nighttime, containing multiple cars, traffic lights, and traffic signs \\ 
5 & 
 depict various city and residential scenes during the daytime with different weather conditions and contain a variety of vehicles, pedestrians, traffic lights, and signs, while the outside examples depict specific limited scenarios with fewer elements., \\ 
6 & depict various traffic scenes, mostly highways at night and during clear weather, with a mix of cars, trucks, buses, traffic lights, and traffic signs\\ 
7 & depict various city streets and highways with clear weather and a moderate amount of traffic, including cars, signs, and occasionally pedestrians, bicycles, and trucks\\ 
8 & depict various urban and residential scenes during different times of day and weather conditions, with a diverse range of vehicles, pedestrians, traffic signs, and traffic lights present\\ 
9 & depict various scenes of city streets and highways with typical traffic conditions and without severe weather conditions \\ 
10 & depict various scenes of city streets and highways with varying weather conditions, traffic, and signage. \\ 
\bottomrule
    \end{tabular}}
    \label{tab:regions_bdd}
\end{table}

\subsection{MMLU Study}

\paragraph{Task.} We rely on the MMLU dataset \cite{hendrycks2020measuring} where participants are shown a question, four possible answers and have to pick the best answer (see screenshots in Figure below for examples). We use ChatGPT, also known as GPT 3.5 turbo as our AI model \cite{openai2020chatgpt}. We obtain the predictions of ChatGPT on the MMLU dataset following the approach in the official repo of MMLU \footnote{\url{https://github.com/hendrycks/test}}. We also ask ChatGPT to explain it's answer by using the prompt "Please explain your answer in one sentence". Both the AI answer and the explanation are shown. 

ChatGPT obtains an accuracy of 69\% during our evaluation and we restrict our attention to specific subjects within the MMLU dataset. Specifically, we sample 5 subjects (out of the 57 in MMLU) where ChatGPT has significantly better perform performance than average, 5 where it's significantly worse, and 4 subjects where performance is similar to average performance. These subjects are listed here:
\begin{quote}
high school government and politics,
 marketing, 
 high school psychology,
 logical fallacies,
 sociology,
 public relations,
 high school computer science,
 anatomy,
 business ethics,
 elementary mathematics,
 high school statistics,
 machine learning,
 moral scenarios,
 global facts
\end{quote}
We sample 150 questions from each subject and additionally sample 150 questions from OpenBookQA dataset \cite{mihaylov2018can} to use as attention checks as human performance on OpenBookQA is 91\%. 

 We run \algname on both the dataset embeddings, metadata (subject name), and an embedding of ChatGPT explanations separately. We find 10 regions based on the metadata and 2 regions based on the ChatGPT explanations. The regions and their descriptions found by our algorithm are reported in Table \ref{tab:regions_mmlu}.

\begin{table}[H]
    \centering
    \caption{Region descriptions found by \algname for the MMLU user study.}
    \resizebox{1\textwidth}{!}{
    \begin{tabular}{cp{0.8\textwidth}}
        \toprule
        Region ID & Description \\
        \midrule
        1 & Related to marketing, including pricing strategies, branding, communication, product classification, market segmentation, advertising, and supply chain management. \\
        2 & Questions related to psychology and neuroscience. \\
        3 & Questions related to global statistics and trends, ranging from military spending to mental health disorders. \\
        4 & Questions inside the region cover a variety of topics in the field of public relations, including ethical frameworks, evaluation models, common tactics, and regulations. \\
        5 & Mathematical and quantitative questions, involving calculations and problem-solving. \\
        6 & Questions related to sociology, including topics such as social class, symbolic interactionism, bureaucracy, and globalization. \\
        7 & Questions and descriptions of logical fallacies and syllogisms. \\
        8 & Questions focus on US politics, government, and history. \\
        9 & Contains questions related to anatomy and physiology. \\
        10 & Various topics including ethics, regulation, consumer rights, and corporate transparency. \\
        11 & Various questions on different topics such as algebra, biology, public relations, and statistics with multiple options to choose from. \\
        \bottomrule
    \end{tabular}}
    \label{tab:regions_mmlu}
\end{table}

\subsection{Screenshots of User Study Interface for BDD}

\begin{figure}[h]
    \centering
        \resizebox{\textwidth}{!}{

    \includegraphics{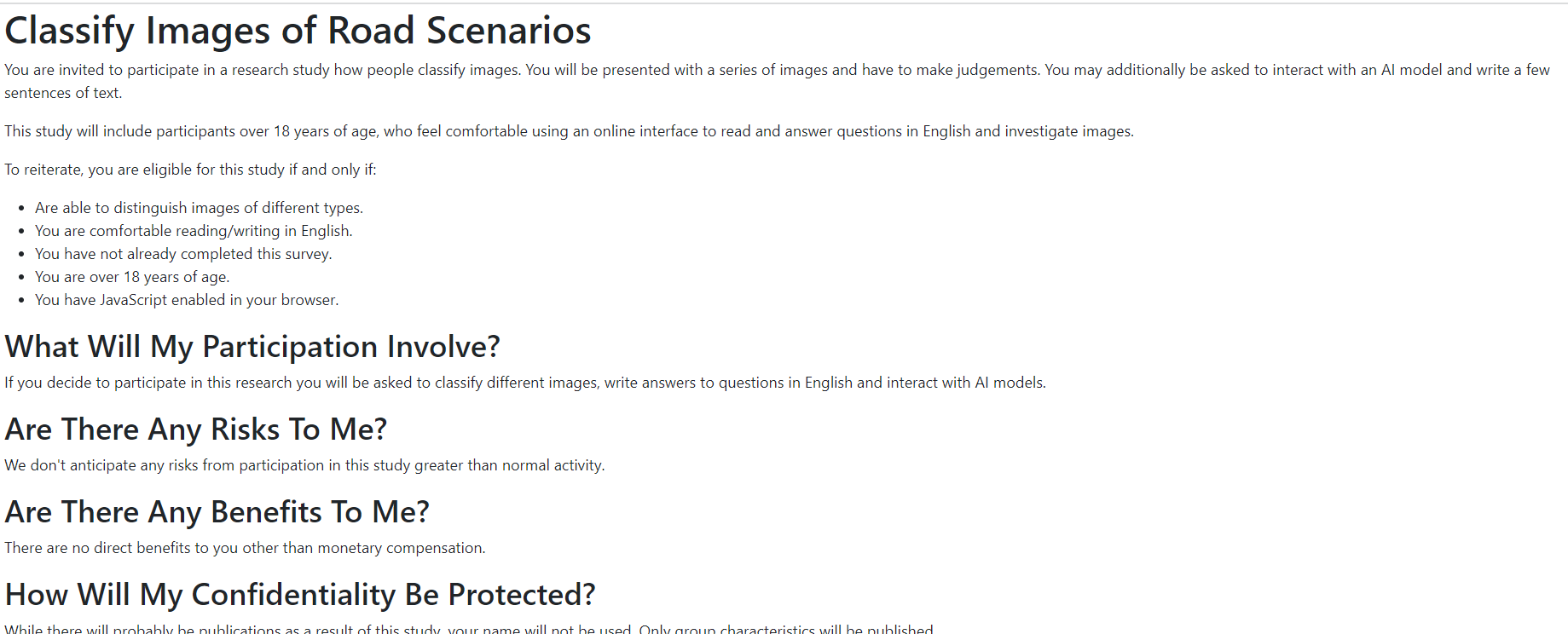}
}
    \caption{Consent Form}
\end{figure}

\begin{figure}[h]
\centering
\resizebox{\textwidth}{!}{
\includegraphics{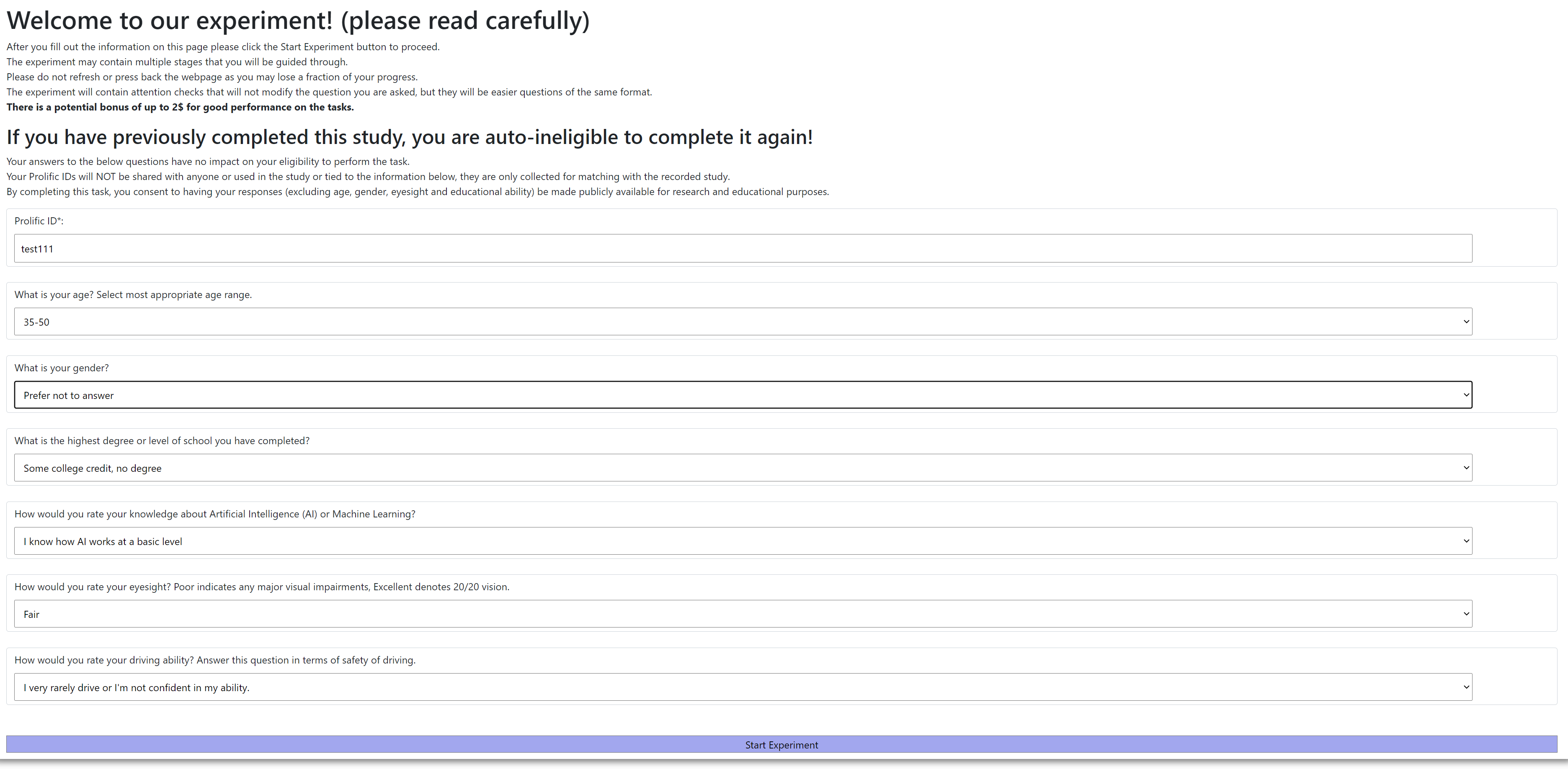}
}
\caption{User Information Collection}
\end{figure}

\begin{figure}[h]
\centering
\resizebox{\textwidth}{!}{
\includegraphics{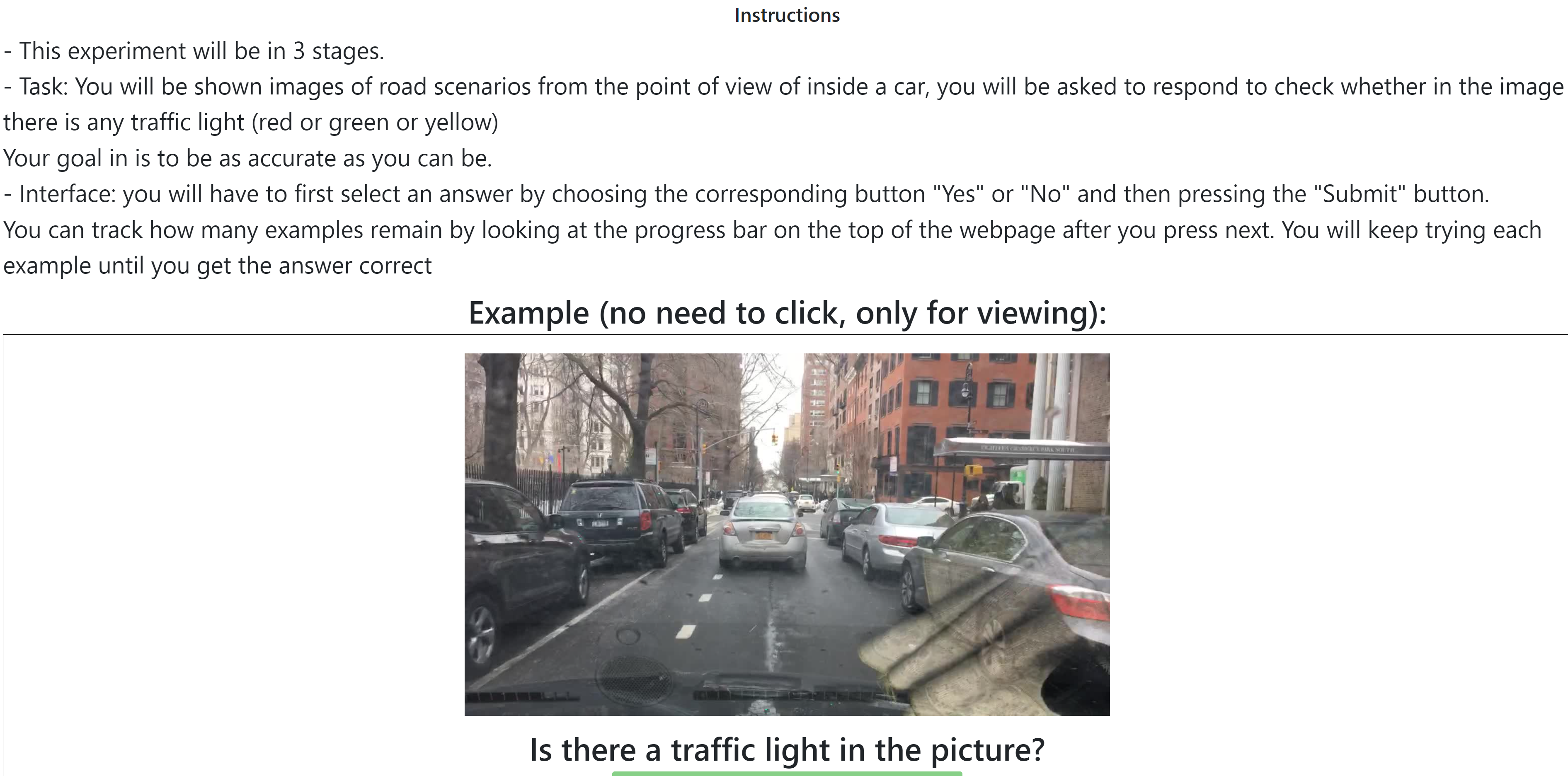}
}
\caption{Practice Task instructions}
\end{figure}

\begin{figure}[h]
\centering
\resizebox{\textwidth}{!}{
\includegraphics{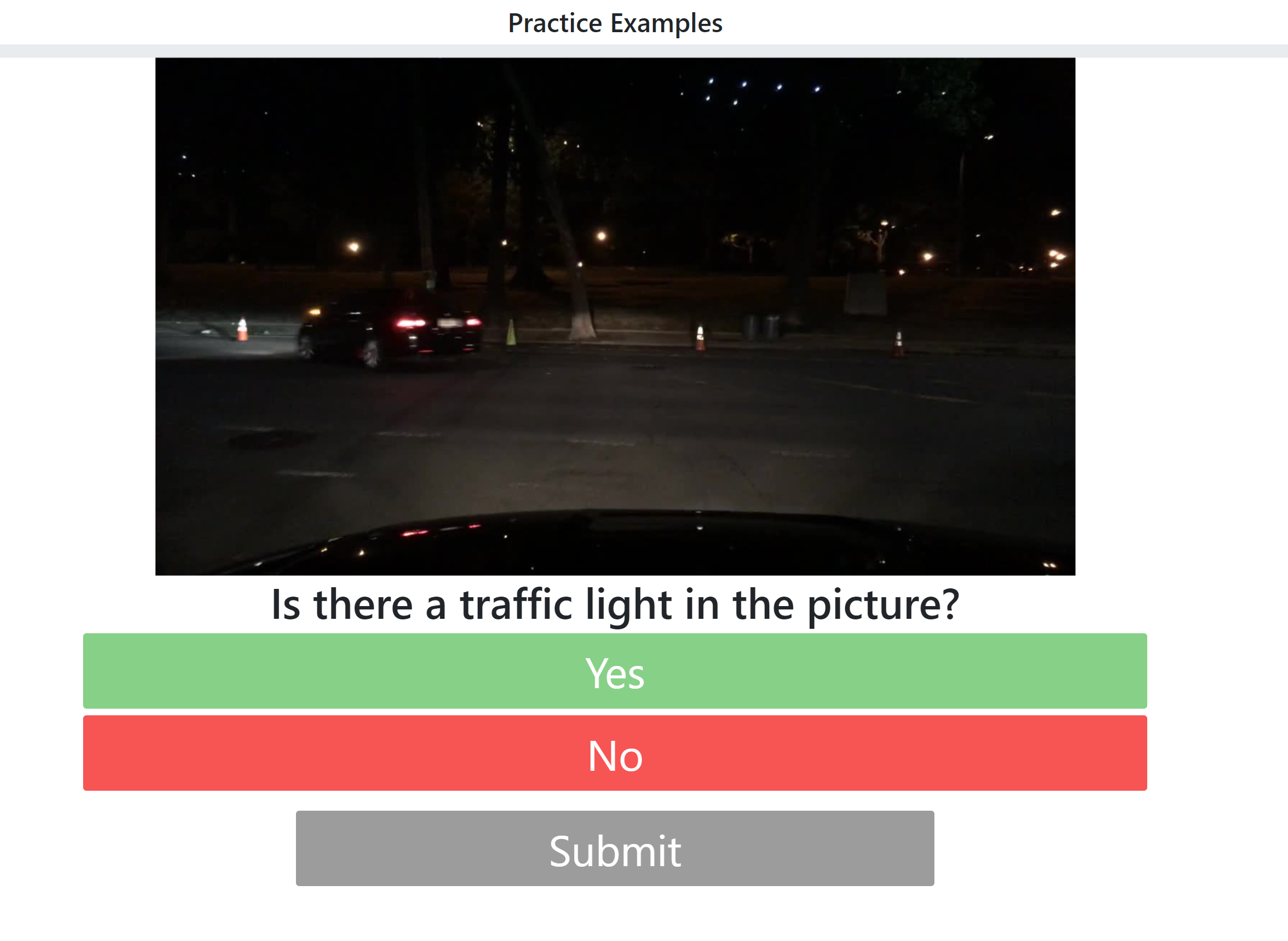}
}
\caption{Prediction without AI interface}
\end{figure}

\begin{figure}[h]
\centering
\resizebox{\textwidth}{!}{
\includegraphics{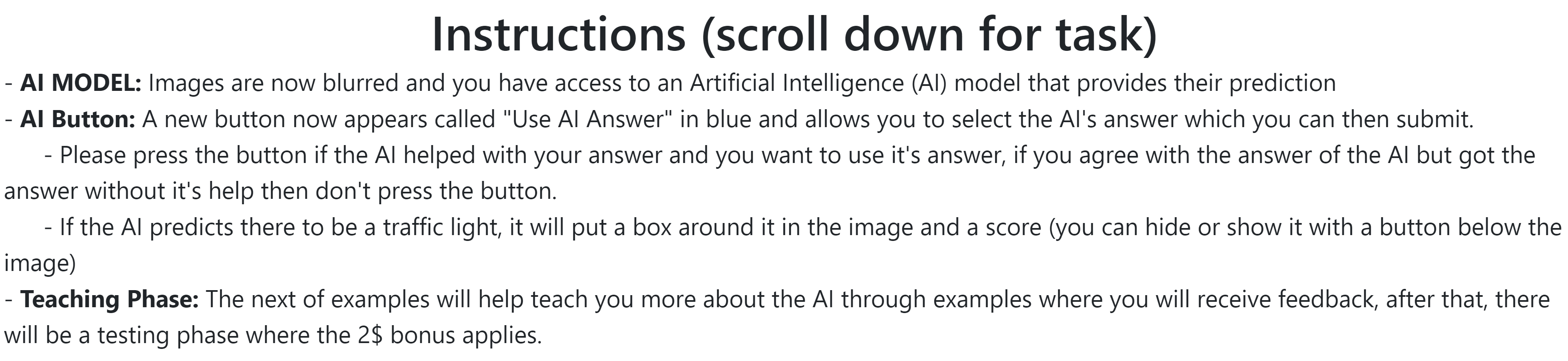}
}
\caption{Instructions for the onboarding phase}
\end{figure}

\begin{figure}[h]
\centering
\resizebox{\textwidth}{!}{
\includegraphics{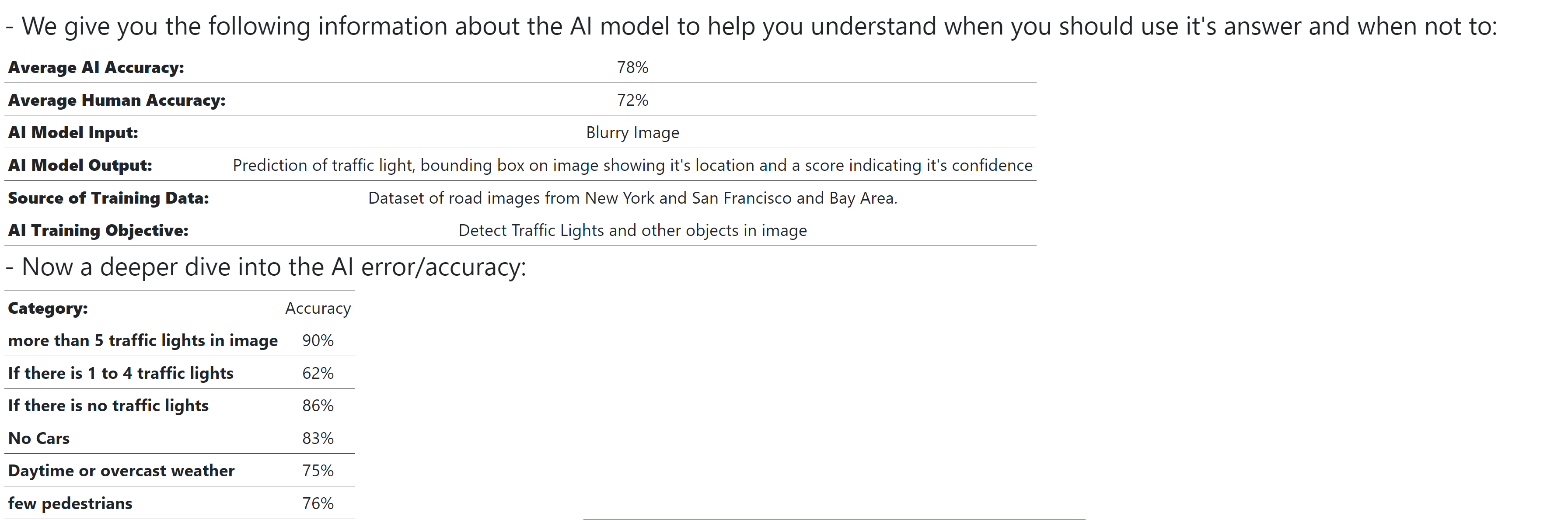}
}
\caption{Model card information shown during onboarding.}
\end{figure}

\begin{figure}[h]
\centering
\resizebox{\textwidth}{!}{
\includegraphics{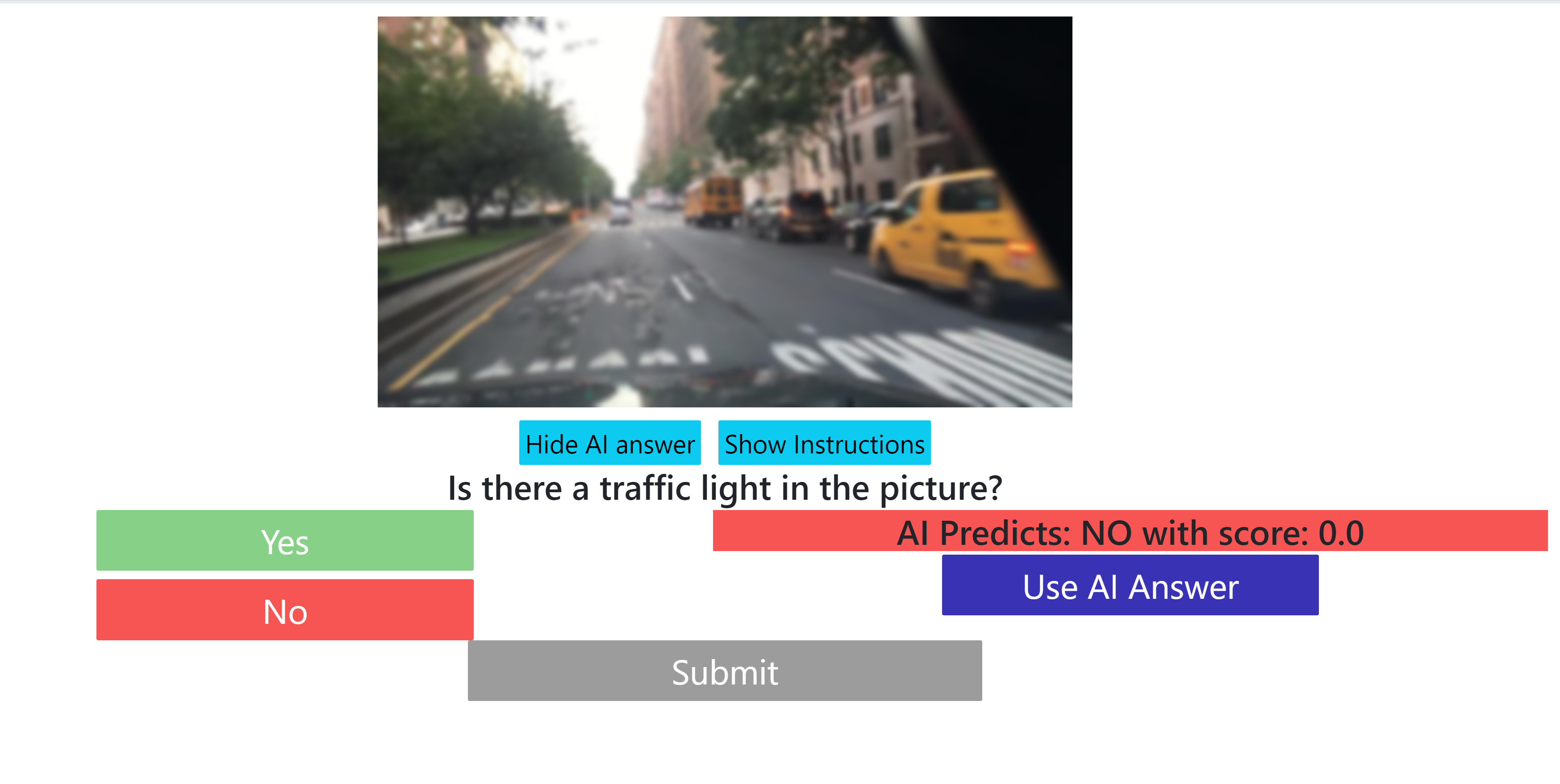}
}
\caption{Prediction with AI interface (AI predicts no traffic light)}
\end{figure}

\begin{figure}[h]
\centering
\resizebox{\textwidth}{!}{
\includegraphics{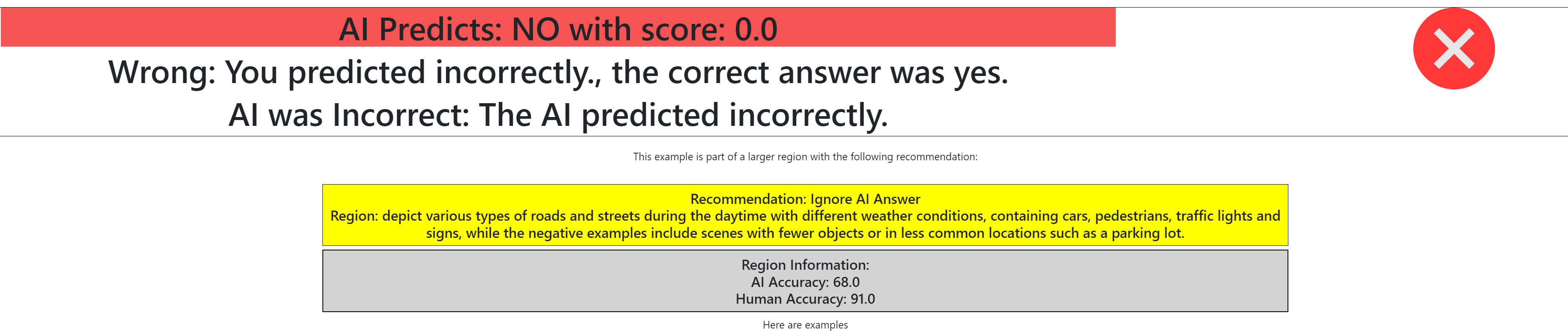}
}
\caption{Feedback shown during onboarding phase after human predicts.}
\end{figure}

\begin{figure}[h]
\centering
\resizebox{\textwidth}{!}{
\includegraphics{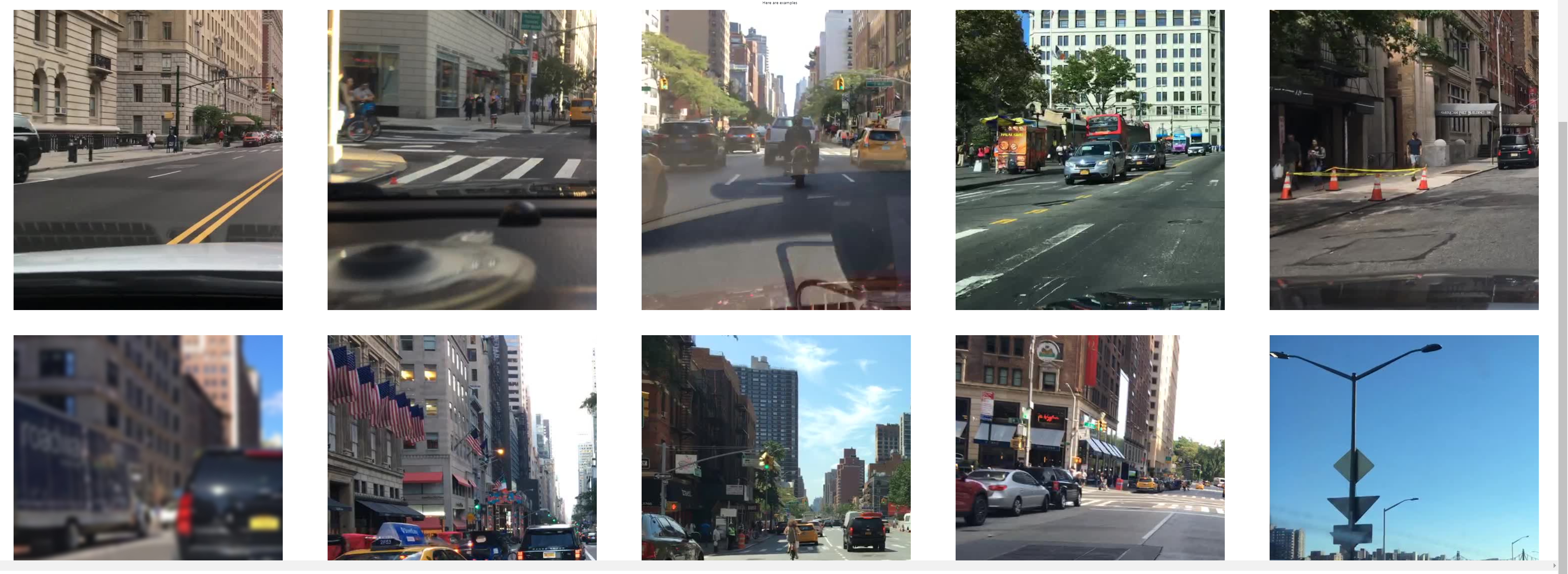}
}
\caption{Feedback shown during onboarding phase after human predicts (sets of examples from region) }
\end{figure}

\begin{figure}[h]
\centering
\resizebox{\textwidth}{!}{
\includegraphics{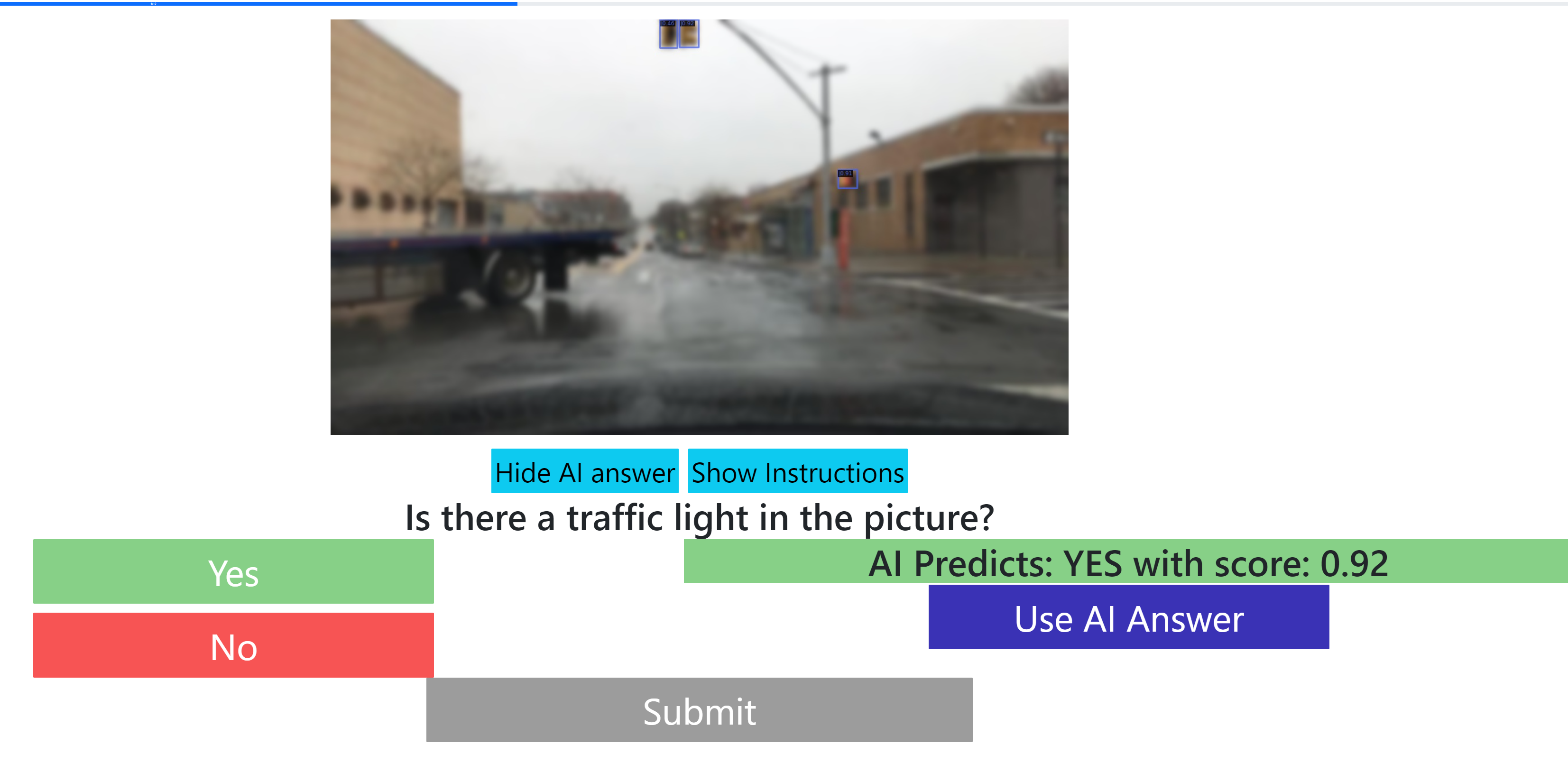}
}
\caption{Prediction with AI interface (AI predicts there is  a traffic light)}
\end{figure}

\begin{figure}[h]
\centering
\resizebox{\textwidth}{!}{
\includegraphics{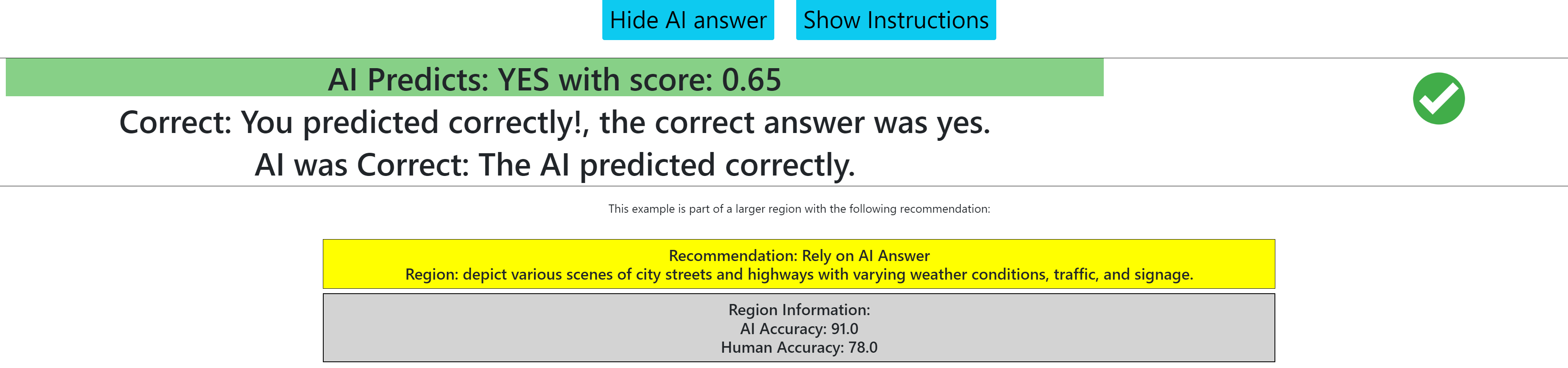}
}
\caption{Feedback shown during onboarding phase after human predicts (correct feedback)}
\end{figure}

\begin{figure}[h]
\centering
\resizebox{\textwidth}{!}{
\includegraphics{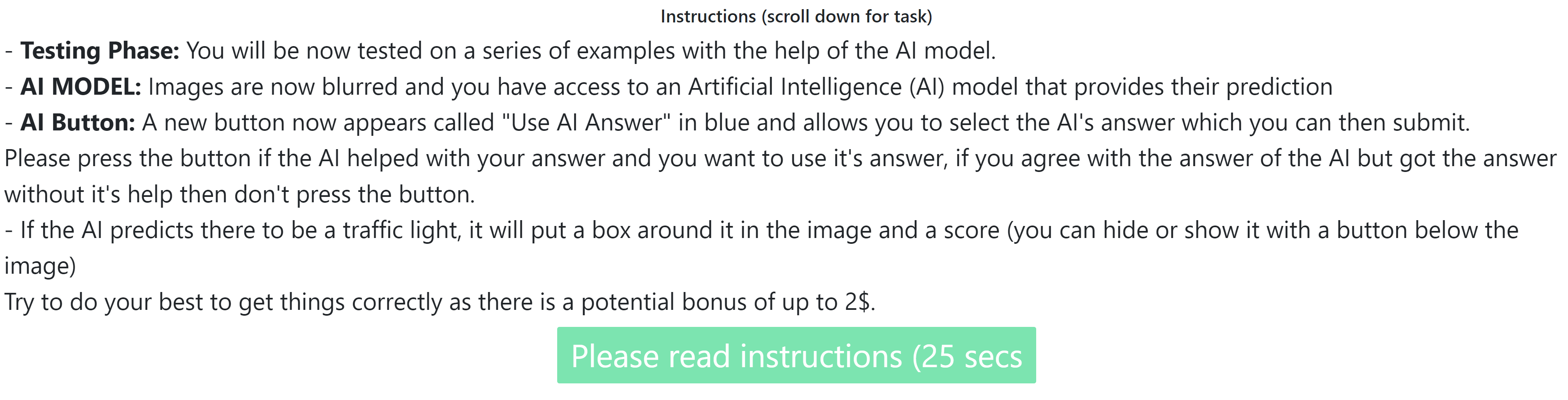}
}
\caption{Testing phase instructions}
\end{figure}

\begin{figure}[h]
\centering
\resizebox{\textwidth}{!}{
\includegraphics{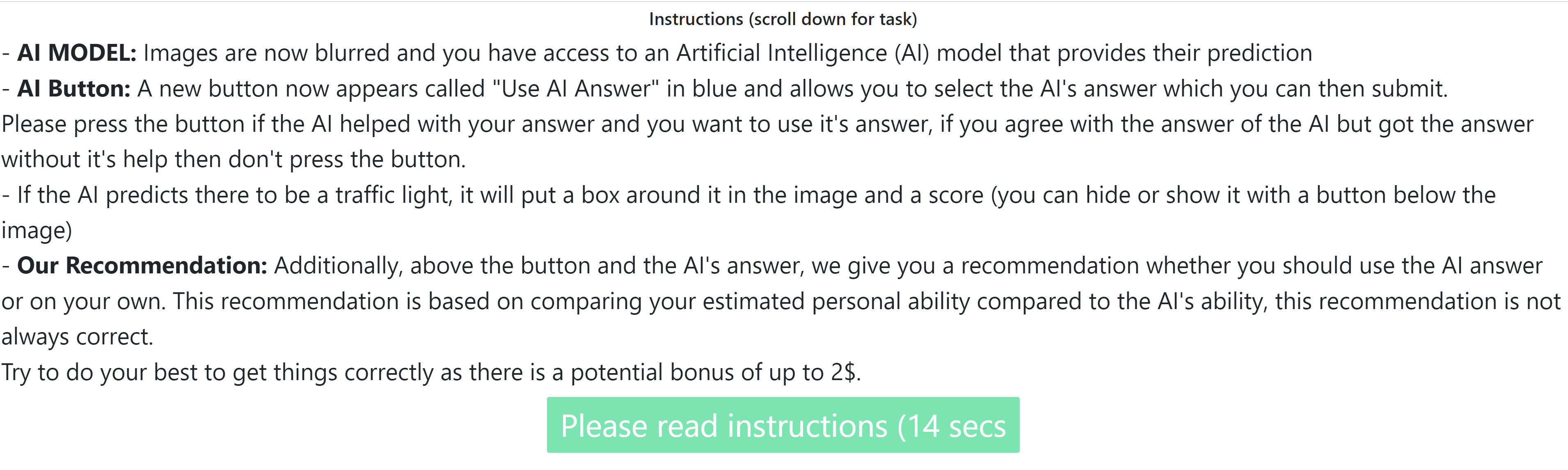}
}
\caption{Testing phase instructions that include AI-integration recommendation}
\end{figure}

\begin{figure}[h]
\centering
\resizebox{\textwidth}{!}{
\includegraphics{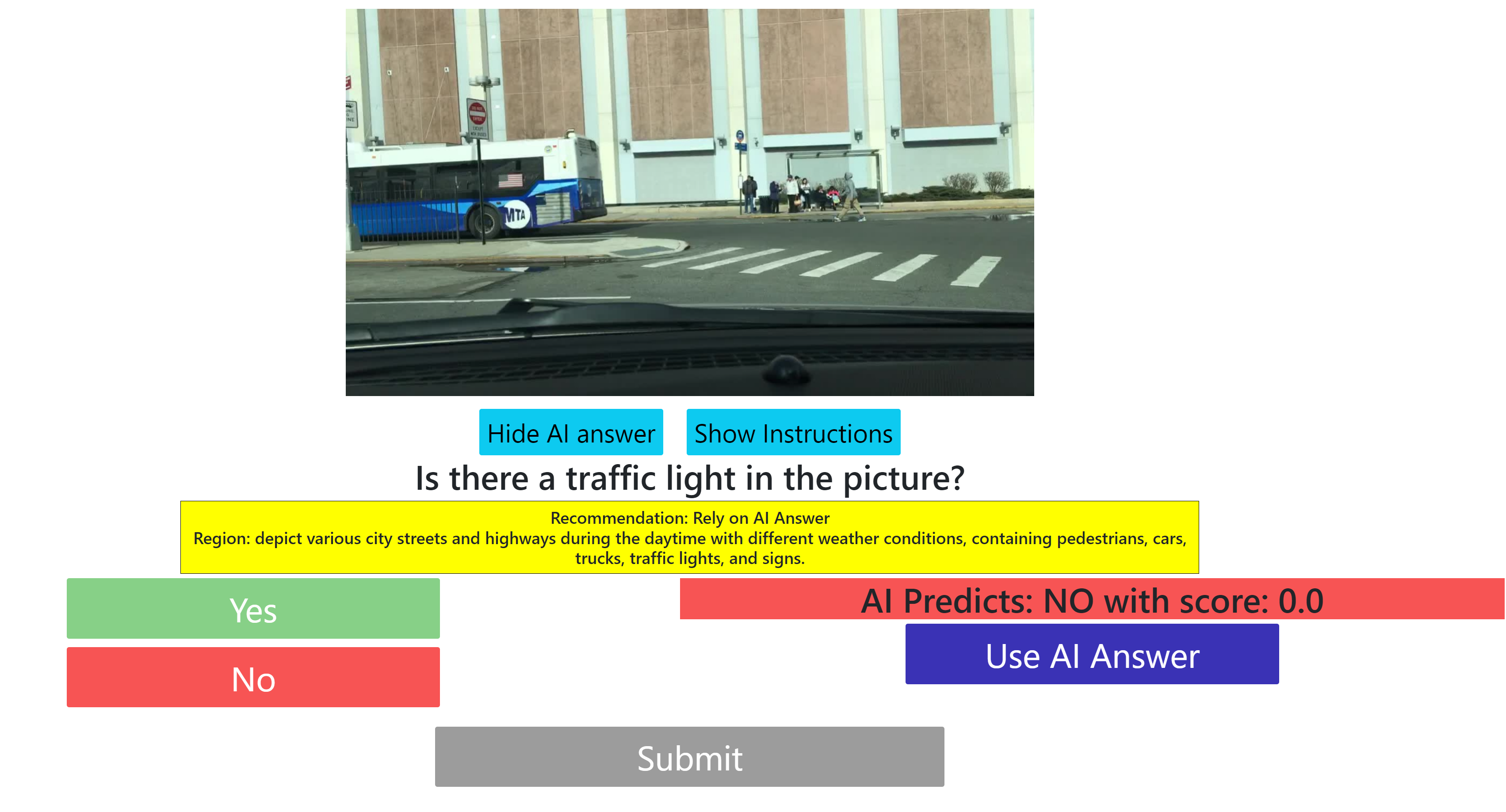}
}
\caption{Prediction interface with AI and with AI-integration recommendations.}
\end{figure}

\clearpage
\newpage
\subsection{Screenshots of User Study Interface for MMLU}

\begin{figure}[h]
\centering
\resizebox{\textwidth}{!}{
\includegraphics{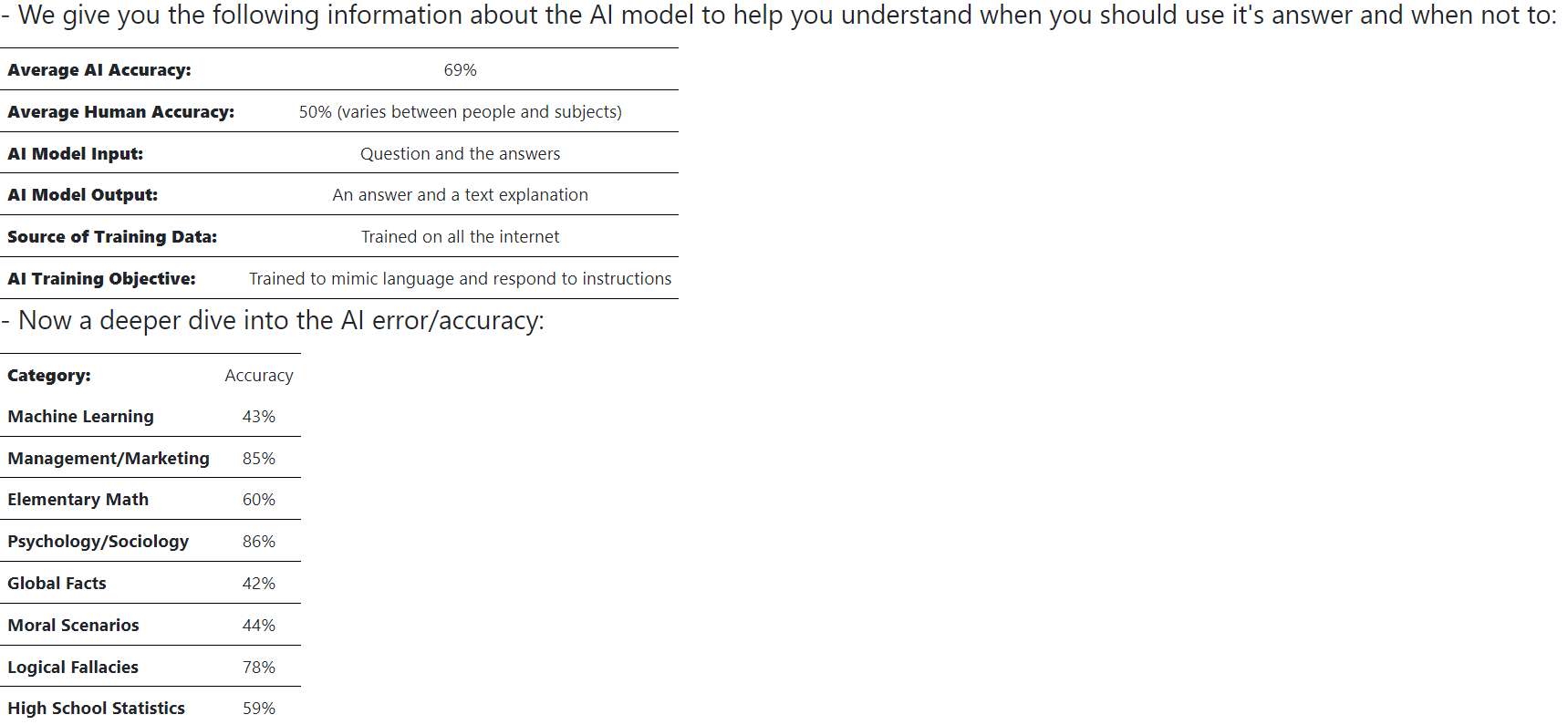}
}
\caption{Model Card for MMLU study}
\end{figure}

\begin{figure}[h]
\centering
\resizebox{\textwidth}{!}{
\includegraphics{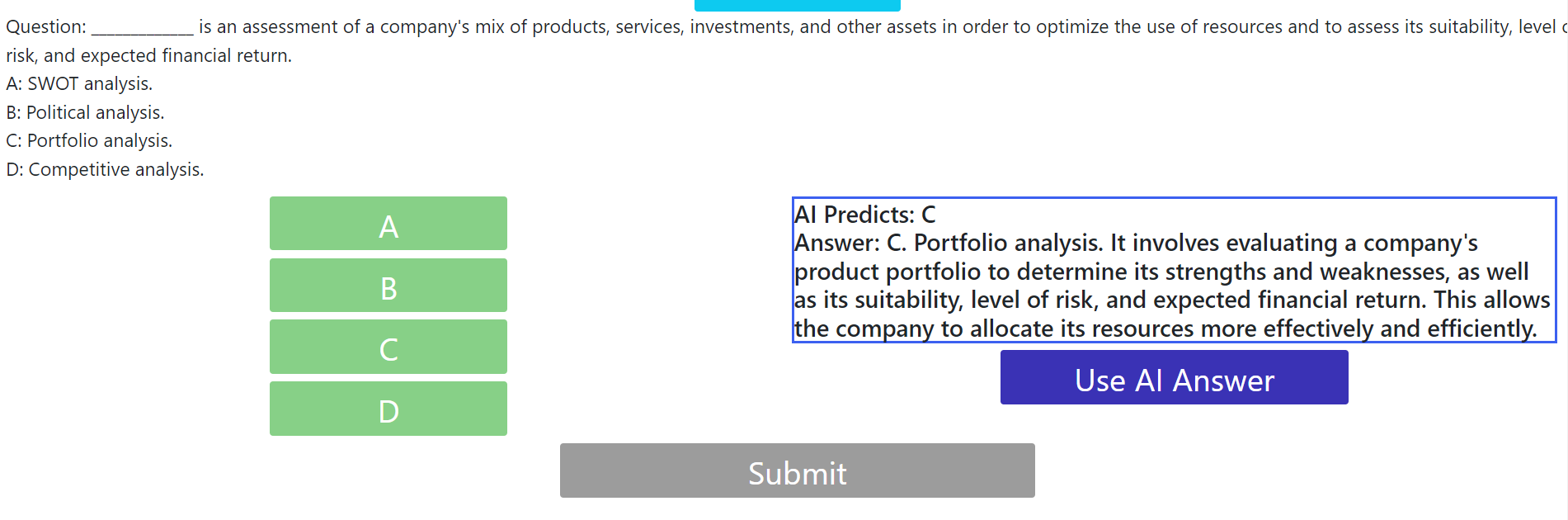}
}
\caption{Prediction Interface for MMLU study}
\end{figure}

\begin{figure}[h]
\centering
\resizebox{\textwidth}{!}{
\includegraphics{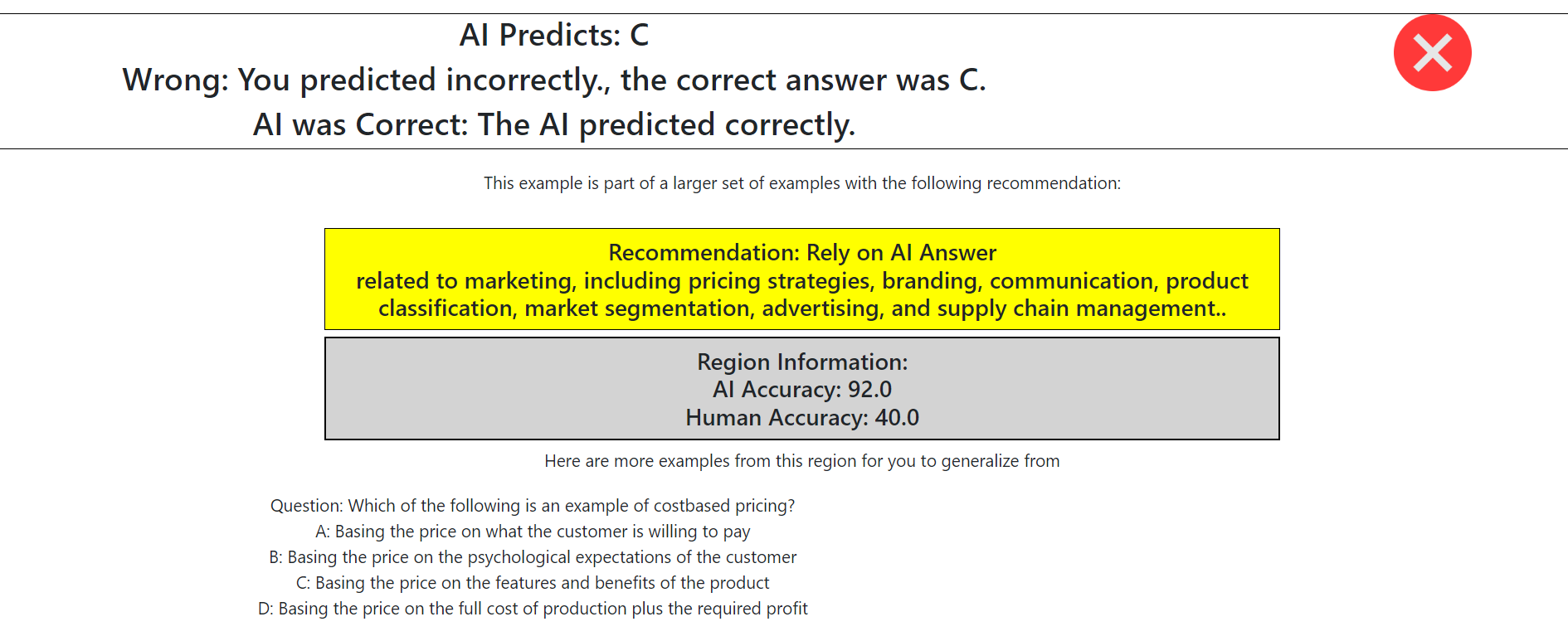}
}
\caption{Feedback shown to user during teaching phase}
\end{figure}

\begin{figure}[h]
\centering
\resizebox{\textwidth}{!}{
\includegraphics{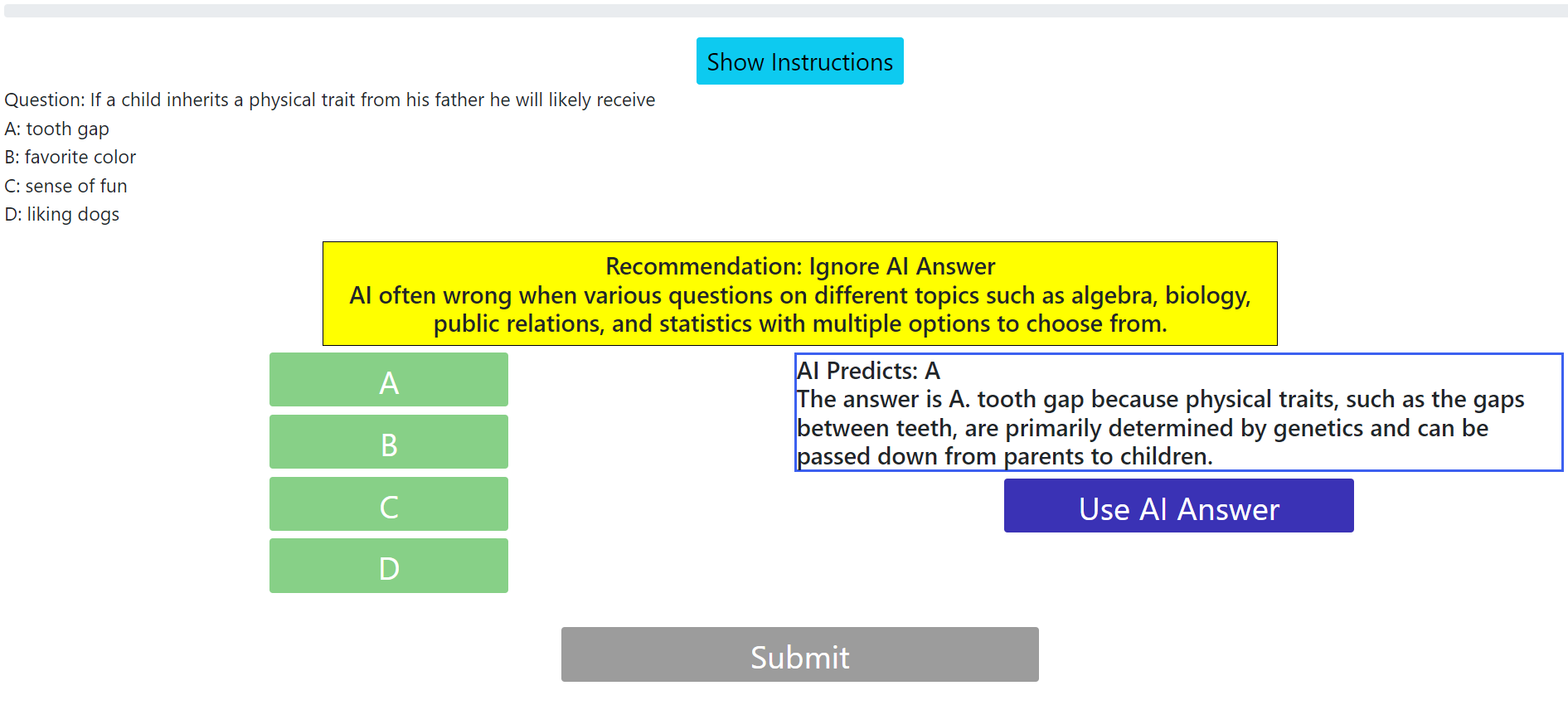}
}
\caption{Prediction Interface for MMLU study with AI-integration recommendations.}
\end{figure}
\end{document}